\documentclass{fairmeta} 
\usepackage{colortbl}
\usepackage{arydshln}

\usepackage[utf8]{inputenc} 
\usepackage[T1]{fontenc}    
\usepackage{url}            
\usepackage{booktabs}       
\usepackage{amsfonts}       
\usepackage{nicefrac}       
\usepackage{microtype}      
\definecolor{bluelink}{RGB}{0,113,188}
\definecolor{greenlink}{RGB}{0,188,113}
\definecolor{PineGreen}{RGB}{0.0, 0.47, 0.44}
\definecolor{Gray}{RGB}{0.5,0.5,0.5}

\definecolor{citecolor}{HTML}{0071bc}
\hypersetup{
    colorlinks=true,%
    citecolor=citecolor,%
    filecolor=citecolor,%
    linkcolor=citecolor,%
    urlcolor=citecolor
}

\usepackage{tabularx}
\usepackage[most]{tcolorbox}
\usepackage{amsmath}
\usepackage{multirow}
\usepackage{array}
\usepackage{caption}
\usepackage{wrapfig}
\usepackage{enumitem}
\usepackage{tikz}
\usepackage{lipsum}
\usepackage{subcaption}
\usepackage{multirow} 
\usepackage{adjustbox}

\usepackage{amssymb}
\renewcommand{\paragraph}[1]{\vspace{1.25mm}\noindent\textbf{#1}}

\newcommand{\finding}[2]{
    \begin{tcolorbox}[
        colback=white!90!gray,     
        colframe=teal!60!black,     
        arc=5pt,                    
        boxsep=5pt,                 
        left=10pt,                  
        right=10pt,                 
        top=2pt,                    
        bottom=2pt,                 
        boxrule=0.8pt,              
        drop shadow=gray!50!white,  
        enhanced jigsaw             
    ]
    \vspace{-0.1cm}
        \paragraph{\textbf{\textit{Finding #1:}}} #2
    \vspace{-0.1cm}
    \end{tcolorbox}
    \vspace{-0.1cm}
}

\usepackage{pgf}
\usepackage{colortbl}
\usepackage{tipa}
\usepackage{tcolorbox}
\usepackage{rotating}
\usepackage[abs]{overpic}
\usepackage{makecell}
\usepackage{longtable}
\usepackage{tocloft}  
\usepackage[english]{babel}
\usepackage{csquotes}
\usepackage{hyperref}

\usepackage[capitalize]{cleveref} 

\newlength\savewidth

\newcommand{\yunyang}[1]{\textcolor{yellow}{}}

\newcommand{\ours}{MetaMorph}
\newcommand{\VPIT}{VPiT}

\title{\ours{}: Multimodal Understanding and Generation via Instruction Tuning}

\author[1,2,*, \dagger]{Shengbang Tong}
\author[1]{David Fan}
\author[1,2,*]{Jiachen Zhu}
\author[3]{Yunyang Xiong}
\author[1]{Xinlei Chen}
\author[1]{Koustuv Sinha}
\author[1]{Michael Rabbat}
\author[1,2]{Yann LeCun}
\author[2]{Saining Xie}
\author[1, \dagger]{Zhuang Liu}

\affiliation[1]{FAIR, Meta}
\affiliation[2]{New York University}
\affiliation[3]{Meta Reality Labs}

\contribution[*]{Work done at Meta}
\contribution[\dagger]{Corresponding authors}

\abstract{
In this work, we propose Visual-Predictive Instruction Tuning (\textbf{\VPIT{}{}})---a simple and effective extension to visual instruction tuning that enables a pretrained LLM to quickly morph into an unified autoregressive model capable of generating both text and visual tokens. \VPIT{} teaches an LLM to predict discrete text tokens and continuous visual tokens from any input sequence of image and text data curated in an instruction-following format.
Our empirical investigation reveals several intriguing properties of \VPIT{}: (1) visual generation ability emerges as a natural byproduct of improved visual understanding, and can be unlocked efficiently with a small amount of  generation data;
(2) 
while we find understanding and generation to be mutually beneficial, understanding data contributes to both capabilities more effectively than generation data.  
Building upon these findings, we train our \textbf{\ours{}} model and achieve competitive performance on both visual understanding and generation. In visual generation, \ours{} can leverage the world knowledge and reasoning abilities gained from LLM pretraining, and overcome common failure modes exhibited by other generation models. 
Our results suggest that LLMs may have strong ``prior'' vision capabilities that can be efficiently adapted to both visual understanding and generation with a relatively simple instruction tuning process.}

\date{\today}

\correspondence{\email{st5087@nyu.edu}, \email{zhuangl@meta.com}}
\project{\href{https://tsb0601.github.io/metamorph}{\sffamily \fontsize{8.8pt}{11pt}\selectfont \texttt{tsb0601.github.io/metamorph}}}

\begin{document}

\maketitle

\section{Introduction} \label{sec: intro}

Multimodal Large Language Models (MLLMs) have advanced considerably in visual understanding, progressing from basic image captioning to complex visual inferences~\citep{alayrac2022flamingo, liu2023visual, dai2024instructblip}.
These models process multimodal inputs---primarily images and language---and generate text tokens. 
Multimodal LLMs often leverage a pretrained vision encoder~\citep{dosovitskiy2020image, radford2021learning}, a pretrained language model~\citep{touvron2023llama2, llama3modelcard}, and align these modalities through connectors such as MLP~\citep{liu2023visual, liu2023improved} or cross-attention modules~\citep{alayrac2022flamingo, dai2024instructblip}. Among MLLM training methods, visual instruction tuning~\citep{liu2023visual} has become widely used~\citep{wang2024qwen2, agrawal2024pixtral}. It treats output embeddings of pretrained vision encoders as continuous-valued ``visual tokens'' and directly feeds them as inputs to pretrained LLMs. 

One benefit of visual instruction tuning is that it is data and compute efficient. A pretrained LLM can be repurposed as a Multimodal LLM by instruction tuning with modest compute and data on the order of millions of image-text question-answer pairs~\citep{tong2024cambrian, li2024llava}. The effectiveness of visual instruction tuning indicates that LLMs already possess a considerable amount of inherent visual knowledge which allows them to efficiently learn and develop visual understanding during the instruction tuning process~\citep{zhou2024lima}. Inspired by this, we investigate whether LLMs can also be finetuned to \textit{generate} visual information with comparable efficiency and effectiveness.

Current attempts toward ``unified'' models---models capable of both multimodal understanding and generation---often treat visual generation as an orthogonal capability to visual understanding. They tend to require substantial changes to the original MLLM architecture and significant multimodal pretraining and/or finetuning.
Designing such methods is challenging, and past research takes different approaches including tokenizing visual inputs into discrete tokens~\citep{wu2024vila, team2024chameleon, liu2024world}, incorporating diffusion objectives~\citep{xie2024show, zhou2024transfusion}, and decoupling vision into separate understanding and generation modes~\citep{wu2024janus}. For example, approaches like LWM~\citep{liu2024world}, Show-o~\citep{xie2024show}, and Chameleon~\citep{team2024chameleon} require billions of image-text pairs~\citep{schuhmann2022laion, gadre2024datacomp} for extensive pretraining and finetuning.

\begin{figure*}[t]
    \vspace{-1.25em}
    \centering
    \hspace{-0.5em}
    \includegraphics[width=\linewidth]{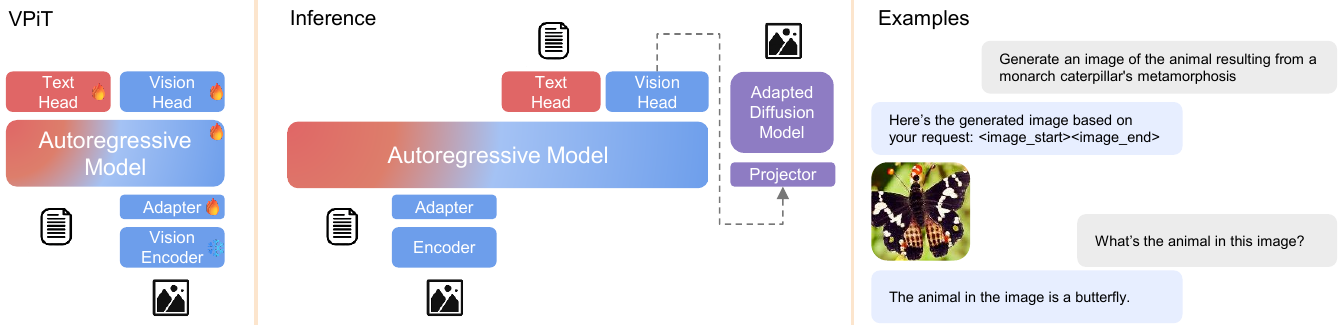}
    \vspace{-0.2em}
    \caption{\small \textbf{\VPIT{} Training, Inference, and Examples of \ours{}.} \textbf{Left}: In Visual-Predictive Instruction Tuning (\VPIT{}), we finetune a pretrained LLM to generate both text and visual tokens using separate text and vision heads. \textbf{Middle}: During inference, the model accepts an arbitrary input sequence of image(s) and text and outputs discrete text tokens and continuous visual tokens. These visual tokens can be visualized via a separately finetuned diffusion model, which is trained to condition on the pretrained vision encoder's output. 
    \textbf{Right}: An example conversation from \ours{} trained with \VPIT{}. Here, the model implicitly solves a visual puzzle in order to generate the visual tokens
    of a butterfly. The conversation continues with new user questions as the model continues to autoregressively process vision and text tokens, independent of the diffusion-based visualization.}
    \label{fig: pipeline}
    \vspace{-1.0em}
\end{figure*}

In this work, we propose \textit{Visual-Predictive Instruction Tuning} (\VPIT{})---a simple extension to visual instruction tuning which builds upon the existing paradigm of passing continuous visual tokens as input to the LLM. \VPIT{} trains an LLM to output \emph{both continuous visual tokens
and discrete text tokens} in the finetuning stage. The model takes pretrained vision encoder embeddings as well as text tokens as input, and outputs a combination of text tokens and continuous visual tokens. To visualize the generated visual tokens, we finetune a diffusion model to map the embeddings back into pixel space (see \cref{fig: pipeline} for an example). This framework allows us to study the synergy between visual understanding, visual generation, and pretrained LLMs, which leads to several intriguing findings outlined below. 

First, we show that the ability to predict visual tokens emerges from understanding visual inputs and requires minimal additional training. Similar to visual instruction tuning, \VPIT{} efficiently and effectively \emph{morphs} an LLM into an ``unified'' model that understands and generates multimodal tokens.
When trained jointly with sufficient visual understanding data, this process requires as little as \textit{200k} additional visual generation data. 

We further establish that the abilities to understand and generate visual tokens are \textit{intrinsically linked} and \textit{asymmetrical}. Specifically, increasing understanding data improves visual understanding (measured by higher VQA scores) and generation performance (measured by lower FID scores). Conversely, increasing generation data enhances generation quality and also contributes to stronger visual understanding---but to a lesser degree. Importantly, our findings highlight an asymmetry in how training each ability impacts the model's overall vision performance: understanding-centric training substantially outperforms generation-centric training in improving both visual understanding and generation.

Building upon these findings, we train a unified model called \emph{\ours{}} to predict multimodal tokens with \VPIT{}{}. We leverage diverse data sources ranging from common visual question answering datasets to pure image and video data without text annotations. \ours{} achieves competitive performance on both visual understanding and visual generation benchmarks. Furthermore, we show this unified modeling approach allows models to leverage the power of LLMs. For instance, \ours{} can extract knowledge from the pretrained LLM when generating visual tokens. More surprisingly, we observe that \ours{} can implicitly perform reasoning steps before generating visual tokens---e.g. when prompted with \textit{``the animal resulting from a monarch caterpillar's metamorphosis''}, \ours{} successfully generates an image of a butterfly (\cref{fig: pipeline}).

Our results suggest that 1) training a unified model with instruction tuning is feasible, and 2) LLMs have strong pre-existing visual capabilities which can be activated using significantly fewer samples compared to extensive pretraining. These insights shed light on the development of mixed-modality models. As the community continues to improve visual understanding in Multimodal LLMs \citep{tong2024cambrian, wang2024qwen2, li2024llava} by advancing base LLMs, instruction tuning techniques, and data, we highlight that these efforts may also implicitly lead to models that are better at visual generation.

\section{Visual-Predictive Instruction Tuning}

Visual instruction tuning as introduced by LLaVA~\citep{liu2023visual} demonstrates that LLMs can be taught to understand visual inputs. This is achieved by finetuning on million-scale data. The success of late-fusion instruction tuning suggests that LLMs may already possess innate visual understanding ability. This ability simply needs to be unlocked through lightweight finetuning. Analogously, we hypothesize that LLMs already possess a degree of innate visual generation ability which just needs to be unlocked with lightweight finetuning. 

Motivated by this, we present Visual-Predictive Instruction Tuning (\VPIT{}, \cref{fig: pipeline})---a simple design which extends existing instruction tuning methods to additionally generate visual tokens rather than text alone. We use the same architecture and next-token prediction paradigm to unlock visual generation capabilities without bells and whistles. We take a pretrained LLM and finetune it to predict both discrete text tokens and continuous visual tokens. The visual tokens can be visualized with an adapted diffusion model. 

\subsection{From Unimodal to Multimodal Next-Token Prediction} \label{sec: multimodal_next_token}

The standard instruction tuning setup consists of an input sequence of conversation rounds~\citep{wei2021finetuned, Taori2023Alpaca}:
$(P_i, R_i)_{i=1}^N$,
where $P_i$ and $R_i$ represent prompts and responses for the $i$-th round of conversation, respectively. The model is trained to generate responses based on the prompt.
\VPIT{}{} adds the following mechanisms to a standard instruction tuning setup to unlock visual understanding and generation.

\paragraph{Tokenizing multimodal data.} 
We extend $P_i$ and $R_i$ to include both text and images. To integrate visual data into a pretrained LLM, we process data closely following visual instruction tuning~\citep{liu2023visual}:

\begin{itemize}
    \item \textbf{Text Data}: Text is tokenized into discrete tokens with a standard tokenizer used by the LLM.  
    \item \textbf{Visual Data}: Images are encoded with a pretrained 
    vision encoder such as SigLIP~\citep{zhai2023sigmoid}. The output is continuous visual tokens which are then interpolated to $m = 64$  tokens. To pass the visual tokens as input to the LLM, we apply a trainable projection layer to align the dimensions with the LLM.
\end{itemize}

\paragraph{Model architecture.}
We take a pretrained LLM and finetune it to process arbitrary sequences of text and visual tokens (detailed next in~\cref{sec: data}). We keep the original LLM head for text prediction, and attach a separate vision head to the LLM for predicting visual tokens,  i.e., the output tokens generated by the vision encoder when processing images. The vision head is a projection layer that projects from the LLM's dimension to the vision encoder's dimension. All response tokens can then be trained and predicted autoregressively, with prompt tokens as context.

Unlike conventional visual instruction tuning, in \VPIT{}, visual tokens are also outputs of the LLM---not just inputs. To make the LLM aware of the presence of visual tokens, we introduce special tokens {\small $\langle \texttt{image\_start} \rangle$} and 
    {\small $\langle \texttt{image\_end} \rangle$} to indicate the boundaries of 
    visual token sequences and when to use the vision head.

\paragraph{Loss functions}. The language head outputs a probability distribution over the vocabulary and is trained with cross-entropy loss for next-token prediction. Visual prediction uses cosine similarity loss between the LLM's predicted visual tokens and those from the vision encoder. Consistent with instruction tuning practices, the model only makes predictions and incurs loss on response tokens.

\subsection{Using Broad Types of Data} \label{sec: data}

Because \VPIT{} enables the model to predict both text and visual tokens in its responses, it allows the use of a broader range of training data. Traditional visual instruction tuning, on the other hand, primarily relies on question-and-answer pairs. The majority of our dataset is publicly available, and we categorize it into three major categories below. This categorization enables us to systematically study the model, as detailed in \cref{sec: teach LLM to generate vision} and \cref{sec: Final Performance}. All data types are formatted as instruction tuning style prompt \& response pairs. See further details in \cref{appendix: data preprocessing}.

\begin{enumerate}
    \item \textbf{Visual Understanding Data}: This category includes data that takes image(s) or video as input and outputs text responses. See~\cref{fig: pipeline} for an example. We use:
    \begin{itemize}
        \item \textbf{ImageQA:} Cambrian-7M~\citep{tong2024cambrian}. The model answers questions based on input image(s). \\
        {
        \small
        $P_i \in \{ \langle \texttt{visual tokens} \rangle, \langle \texttt{text prompt} \rangle \} $
        \\
        $R_i \in \{ \langle \texttt{text response} \rangle \} $
        }
        \item \textbf{VideoQA:} VideoStar~\citep{zohar2024videostar} and ShareVideo~\citep{zhang2024direct}. The model answers questions based on the input video. For videos in VideoQA, we process frames at 1 FPS.\\
        {
        \small
        $P_i \in \{ \langle \texttt{visual tokens} \rangle, \cdots, \langle \texttt{visual tokens} \rangle, \langle \texttt{text prompt} \rangle \} $
        \\
        $R_i \in \{ \langle \texttt{text response} \rangle \} $
        }
    \end{itemize}
    
    \item \textbf{Visual Generation Data}: MetaCLIP~\citep{xu2023demystifying}. The model predicts visual tokens based on an image description. We using at most 5 million pairs. We curate the data into question-answering formats.

     {
    \small
    $P_i \in \{ \langle \texttt{text prompt} \rangle \}$ \\
    $R_i \in \{ \langle \texttt{text response} \rangle, \langle \texttt{visual tokens} \rangle\}$
     }
     
     We prompt the model to generate visual tokens with instructions like \textit{``Generate an image of...''}. The text responses are \textit{``Here is an image based on your request...''}. See~\cref{fig: pipeline} for an example. 
    
    \item \textbf{Other Visual Data}: This category includes data that requires the model to predict visual tokens given interleaved input visual tokens and text tokens. We use:
    
    \begin{itemize}
        \item \textbf{Video Data:} SomethingSomethingV2~\citep{goyal2017making} and HowTo100M~\citep{miech2019howto100m}. The model predicts frames in a sequential order. We design different question-answer pairs to probe into the video, such as asking about future frames, past frames, and reordering frames.
        {
        \small
        $P_i \in \{ \langle \texttt{visual tokens} \rangle, \cdots, \langle \texttt{visual tokens} \rangle, \langle \texttt{text prompt} \rangle \} $
        \\
        $R_i \in \{  \langle \texttt{visual tokens} \rangle, \cdots, \langle \texttt{visual tokens} \rangle \} $
        }
        \item \textbf{Visual Thinking Data:} Visualization-of-Thought~\citep{shao2024visual} and VStar~\citep{wu2023vstar}. The model predicts multimodal tokens in its response before addressing problems. For instance, it predicts a zoomed-in view of an image before generating textual responses.
        
        {
        \small
        $P_i \in \{ \langle \texttt{visual tokens} \rangle, \langle \texttt{text prompt} \rangle \} $
        \\
        $R_i \in \{   \langle \texttt{text response} \rangle, \langle \texttt{visual tokens} \rangle,  \langle \texttt{text response} \rangle \} $
        }
        
        In the response, the model will output \textit{``I will think about it visually''}, followed by visual tokens representing a zoomed-in segment of the image, and then proceed to answer the question.
        \item \textbf{Image-to-Image Data:} InstructPix2Pix~\citep{brooks2023instructpix2pix} and Aurora~\citep{krojer2024learning}. The model generates a transformed image conditioned on a text description and an input image. 
        
        {
        \small
        $P_i \in \{ \langle \texttt{visual tokens} \rangle, \langle \texttt{text prompt} \rangle \} $
        \\
        $R_i \in \{ \langle \texttt{visual tokens} \rangle \} $
        }
        
    \end{itemize}
\end{enumerate} 

\subsection{Mapping Tokens to Images through Diffusion} \label{sec: diffusion visualization}

Because models trained with \VPIT{}{} learn to predict continuous visual tokens, we need to map the predicted tokens back into pixel space. We leverage the concept of a ``Diffusion Autoencoder''~\citep{bordes2022high,preechakul2022diffusion, pan2023kosmos, koh2024generating, li2024return} in which the diffusion model can be adapted to condition on image embeddings rather than text embeddings.  Specifically, we finetune an existing diffusion model to condition on outputs from the vision encoder using held-out training data. 

At inference time, if the tag token {\small $\langle \texttt{image\_start} \rangle$} is generated, the model begins outputting visual tokens until {\small $\langle \texttt{image\_end} \rangle$}. We then plug the generated visual tokens into the diffusion model to visualize the prediction in pixel space. We use standard latent diffusion model training procedures. Details on the hyperparameters and training setup are provided in \cref{appendix: Diffusion Training}.

\section{Findings on Unlocking Visual Generation} \label{sec: teach LLM to generate vision}

We study the following questions about the effects and synergy of visual understanding and generation, under our \VPIT{} framework:

\begin{tabularx}{\textwidth}{@{}p{0.5cm}X@{}}
\S\ref{sec: generation is triggered} & Can visual generation be unlocked through lightweight tuning, or does it require extensive data? \\[0.25em]
\S\ref{sec: understanding for generation} & Are visual understanding and generation mutually beneficial or orthogonal? \\[0.25em]
\S\ref{sec: asymmetric contribution} & How much does more visual understanding or generation data contribute to understanding and generation quality? \\[0.25em]
\S\ref{sec: cotrain vs twostage} & Which visual understanding tasks correlate the most with generation performance?
\end{tabularx}

\paragraph{Evaluation settings.}  We use 9 ImageQA benchmarks (MMBench, Seed, VStar, MMVP, MMMU, ChartQA, TextVQA, ScienceQA, RealWorldQA) to evaluate different aspects of the model. For image generation, we use the finetuned diffusion model to visualize generated visual tokens and measure FID score (lower is better) and CLIP score (higher is better) on the COCO-30K dataset. Unless otherwise specified, we use LLaMA-3 8B~\citep{llama3modelcard} / SigLIP ViT-SO400M-14@384 ~\citep{zhai2023sigmoid} as the pretrained LLM / vision encoder. We also study the effect of different LLMs in \cref{sec: understanding for generation}. We use instruction tuned versions of the LLMs. We pretrain the adapter between the vision encoder and the LLM following visual instruction tuning~\citep{liu2023visual, liu2023improved}. For experiments in this section, we provide training details in \cref{appendix: training hyperparameters} and include the full results in \cref{appendix: vision prediction ablation}.

\subsection{Visual Generation Can Be Unlocked Efficiently by Joint Training with Visual Understanding} 

\label{sec: generation is triggered}

\begin{figure}[b]
    \centering
    \hspace{-0.5em}
    \begin{minipage}[t]{0.42\linewidth}
        \centering
        \includegraphics[width=\linewidth]{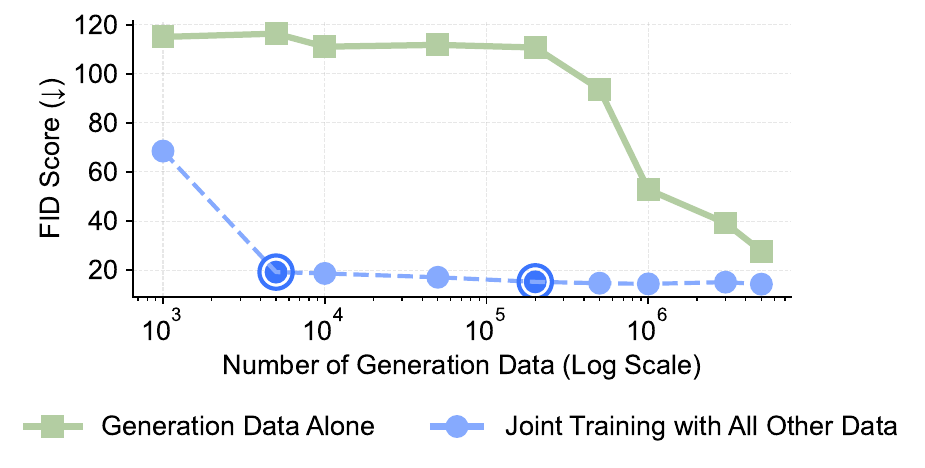}
        \caption{\small \textbf{Generation-only training vs. Joint training with other data.} Training solely on generation data results in inferior performance. Joint training with additional data enables visual generation with only 5k generation data and yields high-quality outputs with 200k generation data.}
        \label{fig: ablation on image-text}
    \end{minipage}
    \hfill
    \begin{minipage}[t]{0.57\linewidth}
        \centering
        \includegraphics[width=\linewidth]{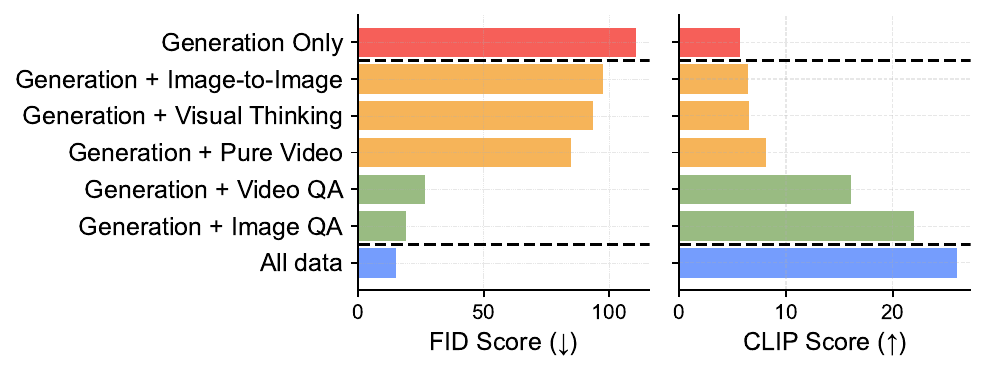}
        \caption{\small \textbf{Impact of different data types on visual generation.} The baseline of training on only visual generation data is \textcolor[HTML]{F5433D}{red}; Joint training with other data is \textcolor[HTML]{F5A83D}{yellow}; Joint training with visual understanding data is \textcolor[HTML]{88B06D}{green}; and all data is \textcolor[HTML]{5D8DFD}{blue}. Joint training with additional data improves the baseline, with \textcolor[HTML]{88B06D}{visual understanding tasks} contributing the most to enhancing visual generation.}
        \label{fig: data type comparison}
    \end{minipage}
\end{figure}

We start by investigating the number of image-text samples required to teach a language model to generate high-quality visual tokens. To this end, we randomly sample \{1k, 5k, 10k, 50k, 200k, 1M, 3M, 5M\} image-text pairs from our generation data (MetaCLIP dataset~\citep{xu2023demystifying}).  We explore two settings: (1) finetuning the LLM using only visual generation data, and (2) joint training visual generation with visual understanding and the rest of data types described in \cref{sec: data}. 

In \cref{fig: ablation on image-text}, we see that training solely on visual generation performs significantly worse than joint training with all other data. With over 3 million image-text pairs, the model struggles to generate high-quality visual images ($\sim$ 40 FID score), and performance remains inferior to joint training with 5 million pairs. This suggests that training solely on visual generation data is significantly less sample efficient. This finding aligns with a prior study~\citep{zhang2023pre} which also suggests that LLMs cannot be easily tuned to generate visual tokens when trained with only generation data. In contrast, joint training with other datasets substantially improves generation performance. The model generates effective visual tokens with just \emph{5k} generation data, and performance stabilizes around \emph{200k} samples. This indicates that visual generation is not an orthogonal capability but rather an ability that benefits from other tasks and emerges more effectively with joint training.

To better understand how each type of data contributes to visual generation, we conduct a controlled experiment using 200k visual generation data, joint training individually with each data type defined in \cref{sec: data}. We also compare them with training all the data together. We show results in \cref{fig: data type comparison}. While all data types enhance the model's visual generation, the degree of improvement varies. Visual understanding data, such as ImageQA and VideoQA, significantly boost the model's visual generation capabilities, even when the amount of generation data is kept constant at 200k. This indicates a strong link between the ability to understand visual content and generate visual tokens. Additionally, combining all data types in training further improves performance, suggesting that the benefits from different data types can be additive.

\finding{1}{The ability to generate visual tokens can be unlocked with significantly less generation data when the model is jointly trained with visual understanding data, in contrast to training only on generation data.} 

\subsection{Visual Understanding and Generation are Mutually Beneficial} \label{sec: understanding for generation}
\begin{figure}[t]
    \centering
    \begin{minipage}{0.48\textwidth}
        \centering
        \includegraphics[width=\linewidth]{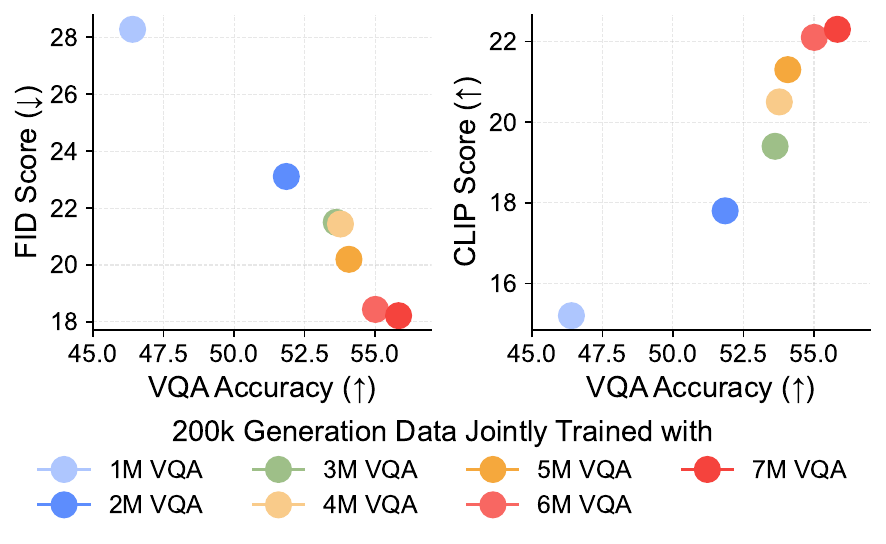}
        \vspace{-1.2em}   
        \caption{\small \textbf{VQA Performance vs. Generation Performance with generation data controlled at 200k.} Increasing understanding data improves VQA and generation performance.}
        \label{fig: vqa vs generation}
    \end{minipage}
    \hfill
    \begin{minipage}{0.49\textwidth}
        \centering
        \includegraphics[width=\linewidth]{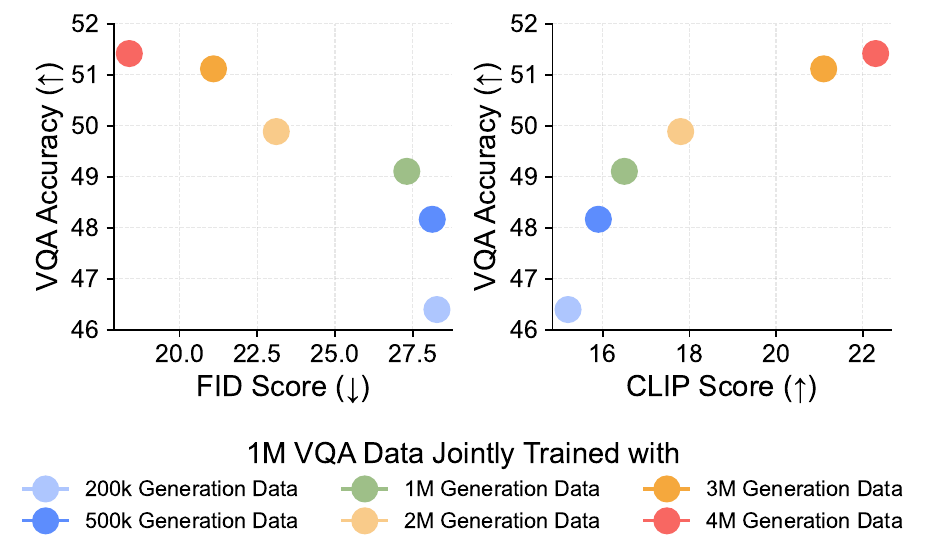}
        \vspace{-1.2em}
        \caption{\small \textbf{Generation Performance vs. VQA Performance with VQA data controlled at 1M.} Increasing generation data improves generation and VQA performance.}
        \label{fig: generation vs vqa}
    \end{minipage}
\end{figure}
\paragraph{More understanding data leads to better understanding and generation.}
Building upon findings from the previous subsection, we perform a controlled experiment to investigate how visual understanding ability correlates with visual generation ability. We ablate our model using a fixed set of 200k generation data while varying VQA data from 1M to 7M samples from Cambrian-7M to develop different levels of visual understanding. The results presented in \cref{fig: vqa vs generation} indicate that stronger VQA ability correlates with better generation performance. 

\paragraph{More generation data leads to better understanding and generation.}
Here, we investigate the reverse direction: does enhancing the model’s visual generation capability also relate to higher VQA performance? To explore this, we conduct a controlled experiment using 1M fixed VQA samples as the baseline for understanding. We then vary the amount of generation data (\{200k, 500k, 1M, 2M, 3M, 4M\}) to adjust generation capacity while joint training with the fixed 1M VQA data. We present results in \cref{fig: generation vs vqa}. Within the 1M VQA setting, stronger generation ability is correlated with improved VQA performance. This implies that increasing the amount of generation data not only enhances generation but also positively impacts VQA performance.

\paragraph{This synergy scales across different LLMs.} 
We examine whether the findings transfer across various LLM backbones. Using a data composition of 7M VQA samples and 1M generation data, we train \VPIT{} on LLaMA-3 8B, LLaMA-3.1 8B, and LLaMA-3 70B. \cref{fig: ablation on LLMs} shows the scaling behavior across different LLMs.

\finding{2}{Visual understanding and generation are synergistic. Increasing data for either capability enhances both simultaneously.}

\subsection{Understanding Data Contributes More} \label{sec: asymmetric contribution}

\begin{figure}[t]
    \centering
    \begin{minipage}[t]{0.48\textwidth}
        \centering
        \includegraphics[width=\linewidth]{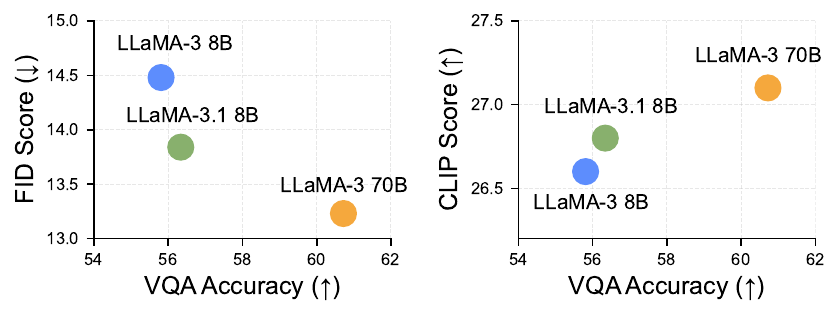}
        \vspace{-1.5em}
        \caption{\small\textbf{Comparison between different language backbones.} We jointly train 7M VQA and 1M Generation data on different language backbones (LLaMA-3 8B, LLaMA-3.1 8B, LLaMA-3 70B). We observe that the synergy between understanding and generation transfer across LLMs.}
        \label{fig: ablation on LLMs}
    \end{minipage}\hfill
    \begin{minipage}[t]{0.48\textwidth}
        \centering
        \includegraphics[width=\linewidth]{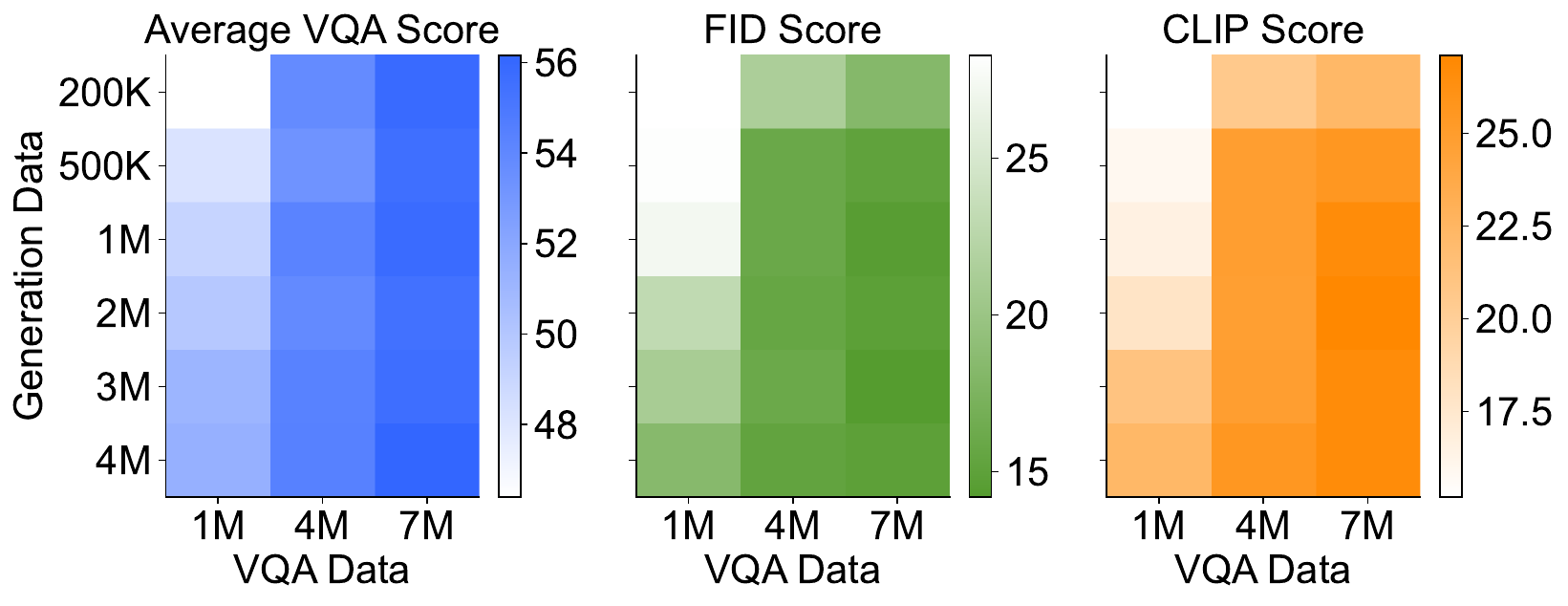}
        \vspace{-1.5em}
        \caption{\small\textbf{Heatmap visualization of Average VQA Score, FID Score, and CLIP Score across varying amounts of VQA data and generation data.} Darker colors indicate better performance. Increasing VQA data is more effective for improving both understanding and generation capabilities.}
        \label{fig: comprehensive vqa vs gen}
    \end{minipage}
\end{figure}

We investigate whether understanding and generation data contribute equally. Here, we jointly train different scales of VQA data \{1M, 4M, 7M\} and generation data \{200k, 500k, 1M, 2M, 3M, 4M\}. \cref{fig: comprehensive vqa vs gen} summarizes these findings, with the x-axis representing VQA data, and the y-axis representing generation data. Results are visualized on heatmaps using darker colors for better performance.

The results indicate that increasing VQA data yields the most significant improvements in all three metrics. When VQA data is relatively low (1M), increases in generation data lead to noticeable improvements, as reflected by the gradual darkening in the plot. However, as the VQA data scales up (from 1M to 4M to 7M), the impact of VQA data becomes more pronounced, demonstrated by a sharp color transition in the heatmap. Ultimately, with 7M VQA data, increases in generation data contribute minimally. These results demonstrate the critical role of understanding data in enhancing both understanding and generation performance.

\finding{3}{While increasing data improves performance overall, the impact of visual understanding data is significantly higher than the impact of visual generation data.}

\subsection{Certain Understanding Tasks Correlate More with Generation Performance}\label{sec: cotrain vs twostage}
\begin{figure*}[h]
    \centering
    \hspace{-0.5em}
    \includegraphics[width=1.01\linewidth]{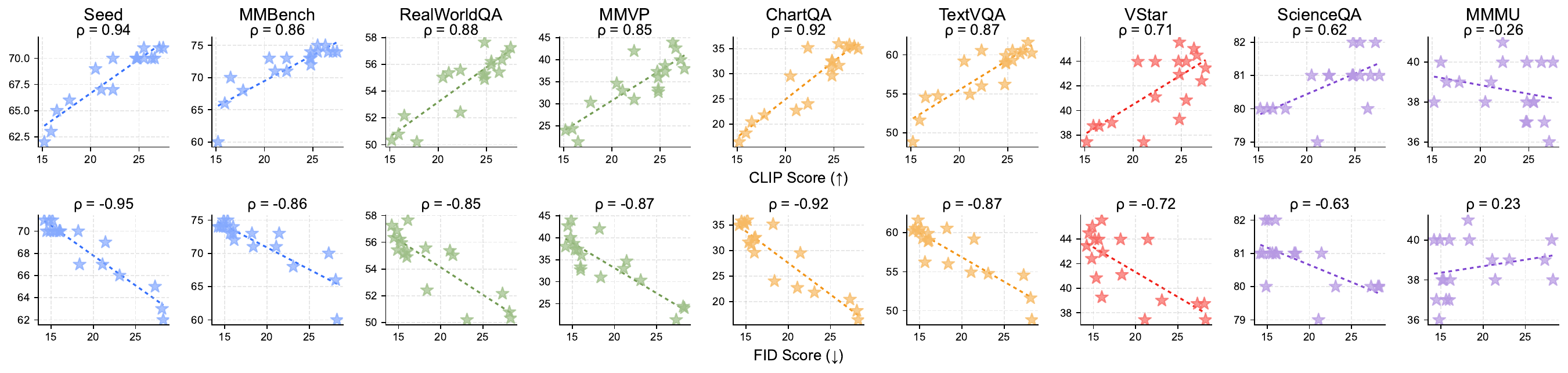}
\vspace{-2em}
    \caption{\small \textbf{Correlation analysis between generation and various understanding benchmarks.} Results are collected by joint training different amounts of VQA data 
    combined with varying quantities of generation data. Each subplot shows the correlation ($\rho$) with a fitted regression line. Stars represent data points. We analyze \textcolor[HTML]{5D8DFD}{General VQA}, \textcolor[HTML]{88B06D}{Vision-Centric VQA}, \textcolor[HTML]{F5A83D}{Text\&Chart VQA}, \textcolor[HTML]{F5433D}{High-Resolution VQA}, and \textcolor[HTML]{9B6FDA}{Knowledge VQA}. For most tasks, generation performance and VQA performance are strongly correlated: higher VQA performance indicates better generation and vice versa. Only knowledge-intensive and high-resolution VQA tasks exhibit weaker correlations with generation performance.} \label{fig: vqa analysis}
\end{figure*}

Given the diverse nature of understanding tasks such as OCR, Vision-Centric tasks, and Knowledge-based tasks, we investigate which tasks most strongly correlate with generation ability. Inspired by Cambrian-1, we categorize VQA tasks into five groups: General, Text\&Chart, High-Resolution, Knowledge, and Vision-Centric VQA.
Using the results from our earlier experiments, which jointly train various VQA data scales with different amounts of generation data, we plot each benchmark’s VQA performance against generation performance in \cref{fig: vqa analysis}. We also calculate the Pearson correlation ($\rho$) between VQA scores and FID/CLIP Scores.

\cref{fig: vqa analysis} shows that General, Vision-Centric, and Text\&Chart VQA tasks strongly correlate with generation performance, each with a Pearson correlation coefficient ($\rho$) above 0.85. High-Resolution VQA exhibits moderate correlation, with $\rho$ around 0.7. In contrast, Knowledge VQA tasks, such as MMMU, show weak correlation with generation performance. These findings suggest that generation ability aligns more closely with the model’s vision capabilities rather than knowledge-specific tasks.

\finding{4}{General, vision-centric, and text understanding VQA tasks exhibit strong correlations with visual generation, whereas knowledge-based VQA tasks do not.} 

\section{\ours{} Model} \label{sec: Final Performance}
Based on the insights in \cref{sec: teach LLM to generate vision}, we train our unified model, \ours{}, based on LLaMA-3.1 8B~\citep{llama3modelcard}, using \VPIT{} with the data curated in \cref{sec: data}. We present our experimental results in three parts: quantitative performance (\cref{sec: quantitative results}), evidence of \ours{} leveraging LLM knowledge in visual generation (\cref{sec: qualitative results}), and implicit reasoning skills in multimodal contexts (\cref{sec: implicit reasoning}).

\begin{table*}[h]
    \centering
    \begin{adjustbox}{width=\linewidth,center}
    \renewcommand{\arraystretch}{1.25}
    \setlength{\tabcolsep}{1.5mm}

    \begin{tabular}{rrccccccccccc}
    \toprule
    
     \multicolumn{2}{c}{} &
     \multicolumn{9}{c}{\textbf{Image QA}} &
     \multicolumn{1}{c}{\textbf{Video QA}} &
     \multicolumn{1}{c}{\textbf{Generation}}  \\
         \cmidrule(lr){3-11}\cmidrule(lr){12-12}\cmidrule(lr){13-13}
      \textbf{Method} & \textbf{Base LLM} &
      \rotatebox{90}{\textbf{{MMBench$^\text{EN}$}}} &
      \rotatebox{90}{\textbf{SEED}} &
      \rotatebox{90}{\textbf{RealworldQA}} &
      \rotatebox{90}{\textbf{MMVP}} &
      \rotatebox{90}{\textbf{SQA}} &
      \rotatebox{90}{\textbf{MMMU}} &
      \rotatebox{90}{\textbf{VStar}} &
      \rotatebox{90}{\textbf{ChartQA}} &
      \rotatebox{90}{\textbf{TextVQA}} &
      \rotatebox{90}{\textbf{MV-Bench}} &
      \rotatebox{90}{\textbf{COCO (FID)}} \\
           \hline
   \textbf{ \color{gray} Visual Understanding Only} &  &   &  &  &  &  &  &  &   &  &  & \\
\color{gray} GPT-4V* &  & \color{gray} 75.8 &  \color{gray} 69.1 & \color{gray}61.4 & \color{gray}50.0 & \color{gray}75.7 &\color{gray} 56.8 & \color{gray}55.0 &\color{gray} 78.5 & \color{gray}78.0 &\color{gray} 43.5 & -    \\
  \color{gray} \textbf{Visual Generation Only} &  &   &  &  &  &  &  &  &   &  &  & \\
\color{gray} Stable Diffusion 1.5$^*$  &  & - & - & - & - & - &  - &  - &  - &  - & - &\color{gray} 9.6  \\
\color{gray} Dalle 2$^*$  &  & - & - & - & - & - &  - &  - &  - &  - & - &\color{gray} 10.4  \\
\color{gray} Imagen$^*$  &  & - & - & - & - & - &  - &  - &  - &  - & - &\color{gray} 7.3  \\
    \hline  
    \textbf{Unified Models} &  &   &  &  &  &  &  &  &   &  &  & \\

    EMU-3$^*$ &  & 58.5 & 68.2 & 57.4 & 36.6$^\dag$ & 89.2 & 31.6 & 51.8$^\dag$ & 68.6 & 64.7 &  - &12.8   \\
    Janus$^*$ & DeepSeek 1.3B & 69.4 & 63.7 & - & - & - & 30.5 & - & - & - &  - & 8.5   \\   
    $\text{VILA-U}_{256}$$^\dag$ & LLaMA-2 7B & 66.6 & 57.1 & 46.6 & 22.0 & 67.1 & 32.2 & 38.7 &  11.4 &  48.3$^*$ & 40.8 & 19.6 \\
     Transfusion$^*$ &  & - & - & - & - & - &  - &  - &  - &  - & - & 6.7\\
     Chameleon-7B$^\dag$
 &  & 35.7 &  27.2 & 19.6 & 0.0 & 50.3 & 28.4 &  37.1 &  0.0 &  0.0 & - &  26.7$^*$ \\
     \rowcolor{blue!5} 
 MetaMorph (ours) & LLaMA-3.1 8B & 75.2 & 71.8 & 58.3 & 48.3 & 83.2 &  41.8 & 44.0 &  37.1 & 60.5 & 48.8 &  11.8  \\
      \bottomrule

    \end{tabular}
    \end{adjustbox}

\captionsetup{font={small}}
 \caption{\small \textbf{Comparison of \ours{} with other unified models.} \ours{} offers competitive performance compared to other leading unified models. Models in \textcolor{gray}{gray} are understanding-only or generation-only. Unified models without a base LLM are trained from scratch. $^*$We use numbers reported in original papers. $^{\dag}$We obtain results using official open-sourced model weights.}
\vspace{-0.25em}
\label{tab:final_table}
\end{table*}

\subsection{Competitive Performance in Understanding and Generation} \label{sec: quantitative results}

We compare \ours{} with other unified models and summarize results in \cref{tab:final_table}. Since these models are trained on different datasets and base LLMs (or pretrained from scratch), an \textit{apples-to-apples} comparison is difficult. Nevertheless, \ours{} demonstrates competitive performance and outperforms other unified models on most benchmarks---even when prior models may have been trained on more data. Compared to models trained from scratch, such as EMU-3~\citep{wang2024emu3} and Chameleon~\citep{team2024chameleon}, \ours{} leverages the strengths of the latest pretrained LLMs and achieves competitive understanding and generation performance. \ours{} highlights that unified models can be developed effectively from pretrained LLMs.

\begin{figure*}[t]
    \centering
    \includegraphics[width=\linewidth]{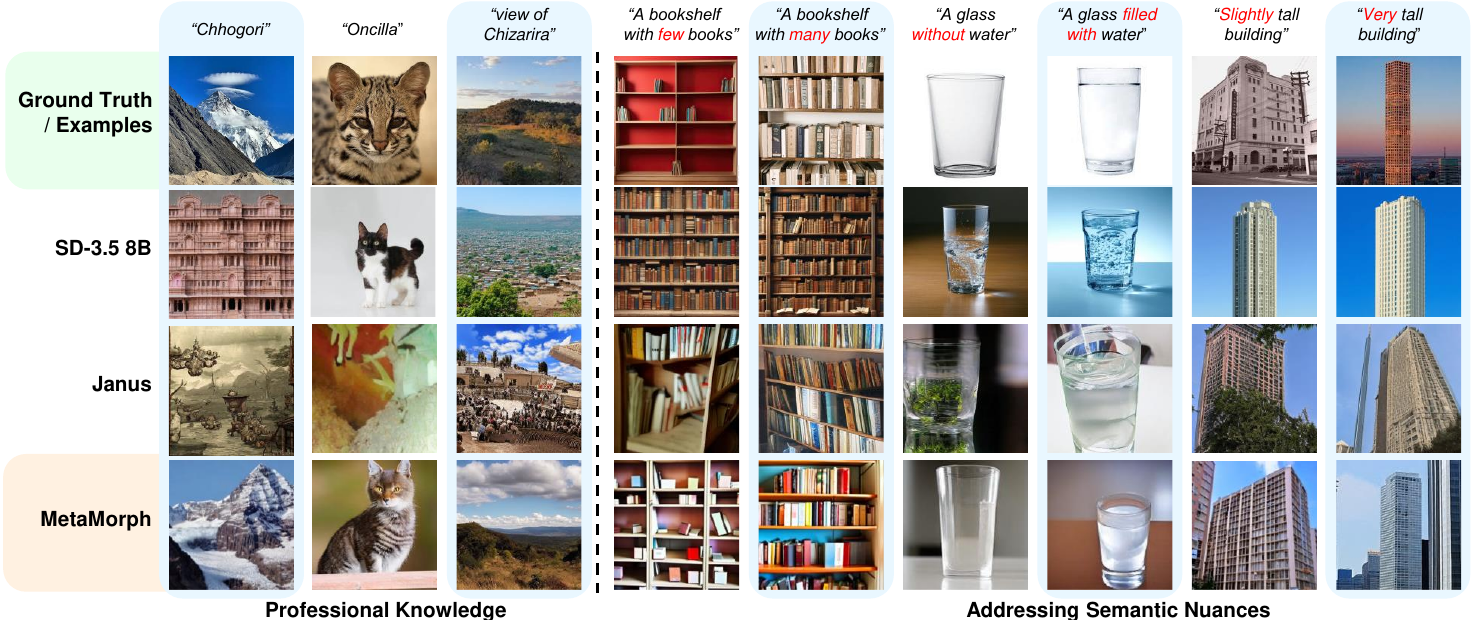}
    \caption{\small \textbf{Examples of \ours{} leveraging LLMs to generate visual tokens.} \textbf{Left}: \ours{} can leverage knowledge from the LLM to generate visual tokens for professional terms that need domain-specific understanding. \textbf{Right}: \ours{} also avoids common mistakes seen in T2I models that condition on text embeddings (e.g., Stable Diffusion-3.5 8B).}
    \label{fig: world knowledge}
    \vspace{-0.2em}
\end{figure*}

\subsection{\ours{} can Leverage LLM Knowledge for Visual Generation } \label{sec: qualitative results}

\ours{} effectively leverages the world knowledge embedded in pre-trained LLMs. We show examples on the left side of \cref{fig: world knowledge}. We prompt the model to generate concepts requiring non-trivial and specialized knowledge. Examples include \textit{``Chhogori''} (the world’s second-highest mountain), \textit{``Oncilla''} (a small wildcat from South America), and \textit{``Chizarira''} (an isolated wilderness area in Zimbabwe). 

\ours{} successfully translates domain-specific knowledge into accurate visual tokens, thereby displaying the ability to \textit{leverage world knowledge from LLMs}. In contrast, the \textit{latest} Text-to-Image (T2I) model, Stable Diffusion-3.5 8B, struggles to generate the correct concept despite producing high-quality images. This issue may stem from the text embedding models it uses—--CLIP~\citep{radford2021learning} and T5~\citep{roberts2019exploring}—--which fail to properly encode these specialized terms~\citep{yuksekgonul2022and}.

On the right side of \cref{fig: world knowledge}, we demonstrate how \ours{} handles common semantic challenges more effectively than text embedding models such as CLIP and T5. These challenges include negation and subjectivity, using prompts with common failure patterns identified in Multimon~\citep{tong2024mass}. \ours{} differentiates semantic nuances such as ``slightly'' versus ``very'', ``few'' versus ``many'', and ``without'' versus ``with'',   which are common failures in existing text-to-image systems.

\subsection{Reasoning in Multimodal Generation} \label{sec: implicit reasoning}

\begin{figure*}[t]
    \centering
    \includegraphics[width=\linewidth]{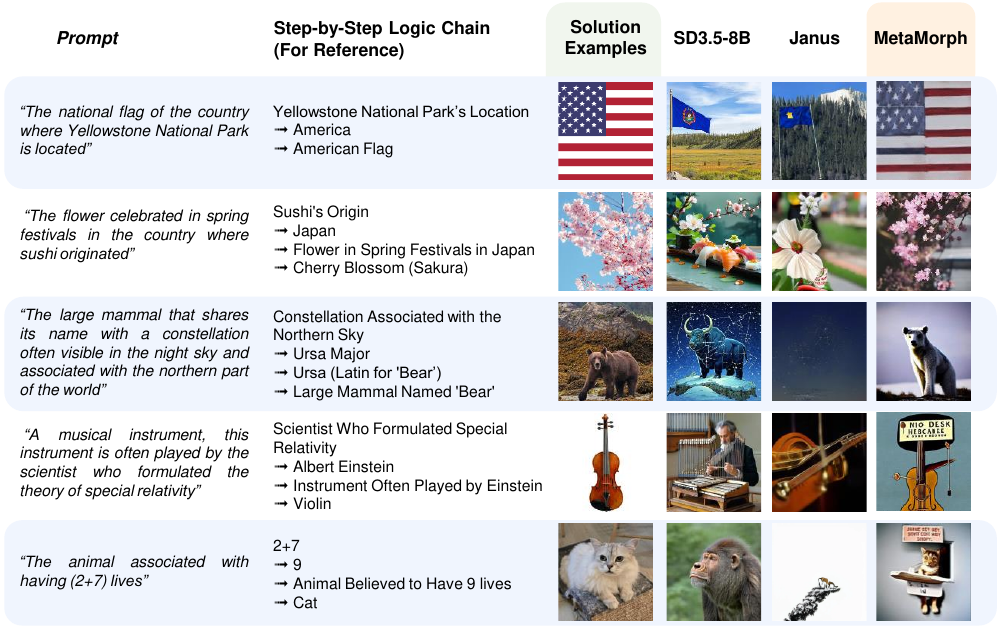}
    \caption{\small \textbf{Examples of \ours{} solving reasoning problems in visual generation.} We design puzzles that require multi-step reasoning. We include reference \colorbox[HTML]{FEF9F3}{\textbf{logic chains}} needed to solve the puzzles, and reference \colorbox[HTML]{F1F6EE}{\textbf{solution examples}}. When prompting each model, we directly feed in the puzzle without any CoT hints or logic chains. \ours{} has the ability to implicitly solve these puzzles and generate the correct image without explicitly creating or processing a logic chain. It demonstrates that the implicit reasoning skills in text-only LLMs can transfer to unified multimodal models.}
    \label{fig: implicit reasoning}
\end{figure*}

In \cref{fig: implicit reasoning}, we present examples where the model generates images in response to puzzle prompts such as \textit{``The national flag of the country where Yellowstone National Park is located''}. For each puzzle, we directly use the prompt \textit{``Generate an image of \{puzzle\}''}, without calling any Chain-of-Thought (CoT)~\citep{wei2022chain} in the prompts.  \ours{} generates the correct image from prompts that require multi-step reasoning. 

For example, when answering the question \textit{``A musical instrument, this instrument is often played by the scientist who formulated the theory of special relativity''}, the model needs to implicitly complete three reasoning steps: it identifies Albert Einstein as the scientist formulated the theory of special relativity, recognizes that his preferred instrument is the violin, and then directly generates correct visual tokens---a violin---without explicitly separating these steps during the generation process. This result implies that \ours{} implicitly solves the puzzle and generates correct visual tokens immediately following the prompt.  These results align with the findings in \textit{Physics of LLMs}~\citep{YXLA2024-gsm1, AllenZhu-icml2024-tutorial}, where the authors suggest that LLMs precompute reasoning graphs before autoregressively generating subsequent tokens. Here, we demonstrate that this capability transfers to the unified multimodal model setting even when decoding visual tokens. 

\section{Related Work} \label{sec: prelimenary}
\paragraph{Instruction tuning and visual instruction tuning.} Instruction tuning~\citep{wei2021finetuned, Taori2023Alpaca} finetunes a pretrained LLM to learn the format and style of interaction. This process helps the model to effectively convey the knowledge and capabilities acquired during pretraining~\citep{zhou2024lima}.
LLaVA~\citep{liu2023visual} extends instruction tuning into the multimodal domain. Since then, different lines of work focus on improving data curation~\citep{chen2023sharegpt4v, laurenccon2024obelics, laurenccon2024matters}, visual representation~\citep{tong2024cambrian, kar2025brave, chen2024far}, and instruction tuning strategies~\citep{gao2024sphinx, liu2024llavanext}. Using only a few million multimodal instruction tuning data, this line of research \citep{liu2024llavanext, tong2024cambrian, li2024llava} has enabled open-source MLLMs to reach performance levels comparable to those of proprietary models~\citep{OpenAI2024gpt4o, Anthropic2024Claude} on a number of benchmarks~\citep{liu2023mmbench, yue2023mmmu, yue2024mmmu} and applications~\citep{zhai2024fine, pan2024autonomous}. 

\paragraph{From Multimodal LLMs to unified models.}
Recent efforts to construct unified models have primarily relied on either extensive pretraining or heavy fine-tuning on billion-scale datasets. Some studies also use continuous embeddings for predicting visual tokens, integrating visual regression losses~\citep{sun2023generative, sun2024generative} or leveraging diffusion-based methods~\citep{dong2023dreamllm}. Other approaches~\citep{lu2022unified, aghajanyan2022cm3, team2024chameleon, wu2024vila, liu2024world, wang2024emu3, lu2024unified} tokenize multimodal data into discrete tokens, which are then trained using autoregressive transformers. Recent research has also explored hybrid strategies that combine autoregressive and diffusion objectives~\citep{zhou2024transfusion, xie2024show}. Different from previous studies, we demonstrate that unified models can be effectively trained in low-data regimes during instruction tuning, while also providing insights into the reciprocal relationship between visual understanding and visual generation.
\section{Discussion}

In this work, we propose \VPIT{}---a simple yet effective extension to visual instruction tuning---that enables LLMs to predict multimodal tokens. 
\VPIT{}{} unlocks the
use of a more diverse range of instruction tuning data than just visual question answering, such as text-to-image and pure image and video data. 
Through controlled experiments, we find that visual generation ability emerges as a natural byproduct of improved visual understanding and requires modest additional generation data. In addition, we find that while visual understanding and generation are mutually beneficial, adding more visual understanding data disproportionately improves overall performance compared to adding more generation data.

Leveraging these insights, we train \ours{} by finetuning LLaMA-3.1 8B with \VPIT{}{}. With a simple training process, \ours{} achieves competitive performance in both visual understanding and generation.
Qualitative evaluation of our model shows that \ours{} can leverage world knowledge and reasoning abilities of the base LLM during visual generation. For example, it can perform multimodal tasks that typically require multiple steps of reasoning,
such as generating images of specialized proper nouns (\textit{``Chhogori''}) or solving visual puzzles (\textit{``generate an image of the animal resulting from a monarch caterpillar's metamorphosis''}).
This indicates that LLMs already possess a degree of ``prior'' visual knowledge which can be activated with only minimal instruction tuning with \VPIT{}{}. 
Overall, LLMs may have a similar representation space as unified and multi-functional models~\citep{huh2024platonic}. We hope the insights from this work inspire more exploration toward developing LLMs for general intelligence.

{
    \small
    \bibliographystyle{ieeenat_fullname}
    \bibliography{main}

\begin{thebibliography}{87}
\providecommand{\natexlab}[1]{#1}
\providecommand{\url}[1]{\texttt{#1}}
\expandafter\ifx\csname urlstyle\endcsname\relax
  \providecommand{\doi}[1]{doi: #1}\else
  \providecommand{\doi}{doi: \begingroup \urlstyle{rm}\Url}\fi

\bibitem[Aghajanyan et~al.(2022)Aghajanyan, Huang, Ross, Karpukhin, Xu, Goyal, Okhonko, Joshi, Ghosh, Lewis, et~al.]{aghajanyan2022cm3}
Armen Aghajanyan, Bernie Huang, Candace Ross, Vladimir Karpukhin, Hu Xu, Naman Goyal, Dmytro Okhonko, Mandar Joshi, Gargi Ghosh, Mike Lewis, et~al.
\newblock Cm3: A causal masked multimodal model of the internet.
\newblock \emph{arXiv preprint arXiv:2201.07520}, 2022.

\bibitem[Agrawal et~al.(2024)Agrawal, Antoniak, Hanna, Chaplot, Chudnovsky, Garg, Gervet, Ghosh, H{\'e}liou, Jacob, et~al.]{agrawal2024pixtral}
Pravesh Agrawal, Szymon Antoniak, Emma~Bou Hanna, Devendra Chaplot, Jessica Chudnovsky, Saurabh Garg, Theophile Gervet, Soham Ghosh, Am{\'e}lie H{\'e}liou, Paul Jacob, et~al.
\newblock Pixtral 12b.
\newblock \emph{arXiv preprint arXiv:2410.07073}, 2024.

\bibitem[AI@Meta(2024)]{llama3modelcard}
AI@Meta.
\newblock Llama 3 model card.
\newblock 2024.

\bibitem[Alayrac et~al.(2022)Alayrac, Donahue, Luc, Miech, Barr, Hasson, Lenc, Mensch, Millican, Reynolds, et~al.]{alayrac2022flamingo}
Jean-Baptiste Alayrac, Jeff Donahue, Pauline Luc, Antoine Miech, Iain Barr, Yana Hasson, Karel Lenc, Arthur Mensch, Katherine Millican, Malcolm Reynolds, et~al.
\newblock Flamingo: a visual language model for few-shot learning.
\newblock In \emph{NeurIPS}, 2022.

\bibitem[{Allen-Zhu}(2024)]{AllenZhu-icml2024-tutorial}
Zeyuan {Allen-Zhu}.
\newblock {ICML 2024 Tutorial: Physics of Language Models}, 2024.
\newblock Project page: \url{https://physics.allen-zhu.com/}.

\bibitem[Anthropic(2024)]{Anthropic2024Claude}
Anthropic.
\newblock Claude, 2024.

\bibitem[Ba et~al.(2016)Ba, Kiros, and Hinton]{ba2016layer}
Jimmy~Lei Ba, Jamie Kiros, and Geoffrey~E. Hinton.
\newblock Layer normalization.
\newblock In \emph{NeurIPS}, 2016.

\bibitem[Bardes et~al.(2024)Bardes, Garrido, Ponce, Chen, Rabbat, LeCun, Assran, and Ballas]{bardes2024revisiting}
Adrien Bardes, Quentin Garrido, Jean Ponce, Xinlei Chen, Michael Rabbat, Yann LeCun, Mahmoud Assran, and Nicolas Ballas.
\newblock Revisiting feature prediction for learning visual representations from video.
\newblock In \emph{TMLR}, 2024.

\bibitem[Bordes et~al.(2022)Bordes, Balestriero, and Vincent]{bordes2022high}
Florian Bordes, Randall Balestriero, and Pascal Vincent.
\newblock High fidelity visualization of what your self-supervised representation knows about.
\newblock In \emph{TMLR}, 2022.

\bibitem[Brooks et~al.(2023)Brooks, Holynski, and Efros]{brooks2023instructpix2pix}
Tim Brooks, Aleksander Holynski, and Alexei~A Efros.
\newblock Instructpix2pix: Learning to follow image editing instructions.
\newblock In \emph{CVPR}, 2023.

\bibitem[Chen et~al.(2023)Chen, Li, Dong, Zhang, He, Wang, Zhao, and Lin]{chen2023sharegpt4v}
Lin Chen, Jisong Li, Xiaoyi Dong, Pan Zhang, Conghui He, Jiaqi Wang, Feng Zhao, and Dahua Lin.
\newblock Sharegpt4v: Improving large multi-modal models with better captions.
\newblock \emph{arXiv preprint arXiv:2311.12793}, 2023.

\bibitem[Chen et~al.(2024{\natexlab{a}})Chen, Wei, Li, Dong, Zhang, Zang, Chen, Duan, Lin, Tang, et~al.]{chen2024sharegpt4video}
Lin Chen, Xilin Wei, Jinsong Li, Xiaoyi Dong, Pan Zhang, Yuhang Zang, Zehui Chen, Haodong Duan, Bin Lin, Zhenyu Tang, et~al.
\newblock Sharegpt4video: Improving video understanding and generation with better captions.
\newblock In \emph{NeurIPS}, 2024{\natexlab{a}}.

\bibitem[Chen et~al.(2024{\natexlab{b}})Chen, Wang, Tian, Ye, Gao, Cui, Tong, Hu, Luo, Ma, et~al.]{chen2024far}
Zhe Chen, Weiyun Wang, Hao Tian, Shenglong Ye, Zhangwei Gao, Erfei Cui, Wenwen Tong, Kongzhi Hu, Jiapeng Luo, Zheng Ma, et~al.
\newblock How far are we to gpt-4v? closing the gap to commercial multimodal models with open-source suites.
\newblock \emph{arXiv preprint arXiv:2404.16821}, 2024{\natexlab{b}}.

\bibitem[Dai et~al.(2024)Dai, Li, Li, Tiong, Zhao, Wang, Li, Fung, and Hoi]{dai2024instructblip}
Wenliang Dai, Junnan Li, Dongxu Li, Anthony Meng~Huat Tiong, Junqi Zhao, Weisheng Wang, Boyang Li, Pascale~N Fung, and Steven Hoi.
\newblock Instructblip: Towards general-purpose vision-language models with instruction tuning.
\newblock In \emph{NeurIPS}, 2024.

\bibitem[Dong et~al.(2024)Dong, Han, Peng, Qi, Ge, Yang, Zhao, Sun, Zhou, Wei, et~al.]{dong2023dreamllm}
Runpei Dong, Chunrui Han, Yuang Peng, Zekun Qi, Zheng Ge, Jinrong Yang, Liang Zhao, Jianjian Sun, Hongyu Zhou, Haoran Wei, et~al.
\newblock Dreamllm: Synergistic multimodal comprehension and creation.
\newblock In \emph{ICLR}, 2024.

\bibitem[Dosovitskiy et~al.(2021)Dosovitskiy, Beyer, Kolesnikov, Weissenborn, Zhai, Unterthiner, Dehghani, Minderer, Heigold, Gelly, et~al.]{dosovitskiy2020image}
Alexey Dosovitskiy, Lucas Beyer, Alexander Kolesnikov, Dirk Weissenborn, Xiaohua Zhai, Thomas Unterthiner, Mostafa Dehghani, Matthias Minderer, Georg Heigold, Sylvain Gelly, et~al.
\newblock An image is worth 16x16 words: Transformers for image recognition at scale.
\newblock In \emph{ICLR}, 2021.

\bibitem[Gadre et~al.(2024)Gadre, Ilharco, Fang, Hayase, Smyrnis, Nguyen, Marten, Wortsman, Ghosh, Zhang, et~al.]{gadre2024datacomp}
Samir~Yitzhak Gadre, Gabriel Ilharco, Alex Fang, Jonathan Hayase, Georgios Smyrnis, Thao Nguyen, Ryan Marten, Mitchell Wortsman, Dhruba Ghosh, Jieyu Zhang, et~al.
\newblock Datacomp: In search of the next generation of multimodal datasets.
\newblock In \emph{NeurIPS}, 2024.

\bibitem[Gao et~al.(2024)Gao, Zhang, Liu, Qiu, Huang, Lin, Zhao, Geng, Lin, Jin, et~al.]{gao2024sphinx}
Peng Gao, Renrui Zhang, Chris Liu, Longtian Qiu, Siyuan Huang, Weifeng Lin, Shitian Zhao, Shijie Geng, Ziyi Lin, Peng Jin, et~al.
\newblock Sphinx-x: Scaling data and parameters for a family of multi-modal large language models.
\newblock \emph{arXiv preprint arXiv:2402.05935}, 2024.

\bibitem[Ge et~al.(2023)Ge, Ge, Zeng, Wang, and Shan]{ge2023planting}
Yuying Ge, Yixiao Ge, Ziyun Zeng, Xintao Wang, and Ying Shan.
\newblock Planting a seed of vision in large language model.
\newblock \emph{arXiv preprint arXiv:2307.08041}, 2023.

\bibitem[Goyal et~al.(2017{\natexlab{a}})Goyal, Ebrahimi~Kahou, Michalski, Materzynska, Westphal, Kim, Haenel, Fruend, Yianilos, Mueller-Freitag, et~al.]{goyal2017something}
Raghav Goyal, Samira Ebrahimi~Kahou, Vincent Michalski, Joanna Materzynska, Susanne Westphal, Heuna Kim, Valentin Haenel, Ingo Fruend, Peter Yianilos, Moritz Mueller-Freitag, et~al.
\newblock The" something something" video database for learning and evaluating visual common sense.
\newblock In \emph{ICCV}, 2017{\natexlab{a}}.

\bibitem[Goyal et~al.(2017{\natexlab{b}})Goyal, Khot, Summers-Stay, Batra, and Parikh]{goyal2017making}
Yash Goyal, Tejas Khot, Douglas Summers-Stay, Dhruv Batra, and Devi Parikh.
\newblock Making the v in vqa matter: Elevating the role of image understanding in visual question answering.
\newblock In \emph{CVPR}, 2017{\natexlab{b}}.

\bibitem[Hendrycks and Gimpel(2016)]{hendrycks2016gaussian}
Dan Hendrycks and Kevin Gimpel.
\newblock Gaussian error linear units (gelus).
\newblock \emph{arXiv preprint arXiv:1606.08415}, 2016.

\bibitem[Hessel et~al.(2021)Hessel, Holtzman, Forbes, Bras, and Choi]{hessel2021clipscore}
Jack Hessel, Ari Holtzman, Maxwell Forbes, Ronan~Le Bras, and Yejin Choi.
\newblock Clipscore: A reference-free evaluation metric for image captioning.
\newblock In \emph{EMNLP}, 2021.

\bibitem[Heusel et~al.(2017)Heusel, Ramsauer, Unterthiner, Nessler, and Hochreiter]{heusel2017gans}
Martin Heusel, Hubert Ramsauer, Thomas Unterthiner, Bernhard Nessler, and Sepp Hochreiter.
\newblock Gans trained by a two time-scale update rule converge to a local nash equilibrium.
\newblock In \emph{NeurIPS}, 2017.

\bibitem[Huh et~al.(2024)Huh, Cheung, Wang, and Isola]{huh2024platonic}
Minyoung Huh, Brian Cheung, Tongzhou Wang, and Phillip Isola.
\newblock The platonic representation hypothesis.
\newblock In \emph{ICML}, 2024.

\bibitem[Kar et~al.(2025)Kar, Tonioni, Poklukar, Kulshrestha, Zamir, and Tombari]{kar2025brave}
O{\u{g}}uzhan~Fatih Kar, Alessio Tonioni, Petra Poklukar, Achin Kulshrestha, Amir Zamir, and Federico Tombari.
\newblock Brave: Broadening the visual encoding of vision-language models.
\newblock In \emph{ECCV}, 2025.

\bibitem[Koh et~al.(2024)Koh, Fried, and Salakhutdinov]{koh2024generating}
Jing~Yu Koh, Daniel Fried, and Russ~R Salakhutdinov.
\newblock Generating images with multimodal language models.
\newblock In \emph{NeurIPS}, 2024.

\bibitem[Krojer et~al.(2024)Krojer, Vattikonda, Lara, Jampani, Portelance, Pal, and Reddy]{krojer2024learning}
Benno Krojer, Dheeraj Vattikonda, Luis Lara, Varun Jampani, Eva Portelance, Christopher Pal, and Siva Reddy.
\newblock Learning action and reasoning-centric image editing from videos and simulations.
\newblock In \emph{NeurIPS}, 2024.

\bibitem[Lauren{\c{c}}on et~al.(2024{\natexlab{a}})Lauren{\c{c}}on, Saulnier, Tronchon, Bekman, Singh, Lozhkov, Wang, Karamcheti, Rush, Kiela, et~al.]{laurenccon2024obelics}
Hugo Lauren{\c{c}}on, Lucile Saulnier, L{\'e}o Tronchon, Stas Bekman, Amanpreet Singh, Anton Lozhkov, Thomas Wang, Siddharth Karamcheti, Alexander Rush, Douwe Kiela, et~al.
\newblock Obelics: An open web-scale filtered dataset of interleaved image-text documents.
\newblock \emph{Advances in Neural Information Processing Systems}, 36, 2024{\natexlab{a}}.

\bibitem[Lauren{\c{c}}on et~al.(2024{\natexlab{b}})Lauren{\c{c}}on, Tronchon, Cord, and Sanh]{laurenccon2024matters}
Hugo Lauren{\c{c}}on, L{\'e}o Tronchon, Matthieu Cord, and Victor Sanh.
\newblock What matters when building vision-language models?
\newblock \emph{arXiv preprint arXiv:2405.02246}, 2024{\natexlab{b}}.

\bibitem[LeCun(2022)]{lecun2022path}
Yann LeCun.
\newblock A path towards autonomous machine intelligence version 0.9. 2, 2022-06-27.
\newblock \emph{Open Review}, 62\penalty0 (1):\penalty0 1--62, 2022.

\bibitem[Li et~al.(2024{\natexlab{a}})Li, Zhang, Guo, Zhang, Li, Zhang, Zhang, Li, Liu, and Li]{li2024llava}
Bo Li, Yuanhan Zhang, Dong Guo, Renrui Zhang, Feng Li, Hao Zhang, Kaichen Zhang, Yanwei Li, Ziwei Liu, and Chunyuan Li.
\newblock Llava-onevision: Easy visual task transfer.
\newblock \emph{arXiv preprint arXiv:2408.03326}, 2024{\natexlab{a}}.

\bibitem[Li et~al.(2024{\natexlab{b}})Li, Wang, He, Li, Wang, Liu, Wang, Xu, Chen, Luo, et~al.]{li2024mvbench}
Kunchang Li, Yali Wang, Yinan He, Yizhuo Li, Yi Wang, Yi Liu, Zun Wang, Jilan Xu, Guo Chen, Ping Luo, et~al.
\newblock Mvbench: A comprehensive multi-modal video understanding benchmark.
\newblock In \emph{CVPR}, 2024{\natexlab{b}}.

\bibitem[Li et~al.(2024{\natexlab{c}})Li, Katabi, and He]{li2024return}
Tianhong Li, Dina Katabi, and Kaiming He.
\newblock Return of unconditional generation: A self-supervised representation generation method.
\newblock In \emph{NeurIPS}, 2024{\natexlab{c}}.

\bibitem[Lin et~al.(2014)Lin, Maire, Belongie, Hays, Perona, Ramanan, Doll{\'a}r, and Zitnick]{lin2014microsoft}
Tsung-Yi Lin, Michael Maire, Serge Belongie, James Hays, Pietro Perona, Deva Ramanan, Piotr Doll{\'a}r, and C~Lawrence Zitnick.
\newblock Microsoft coco: Common objects in context.
\newblock In \emph{ECCV}, 2014.

\bibitem[Liu et~al.(2023)Liu, Li, Wu, and Lee]{liu2023visual}
Haotian Liu, Chunyuan Li, Qingyang Wu, and Yong~Jae Lee.
\newblock Visual instruction tuning.
\newblock In \emph{NeurIPS}, 2023.

\bibitem[Liu et~al.(2024{\natexlab{a}})Liu, Li, Li, and Lee]{liu2023improved}
Haotian Liu, Chunyuan Li, Yuheng Li, and Yong~Jae Lee.
\newblock Improved baselines with visual instruction tuning.
\newblock In \emph{CVPR}, 2024{\natexlab{a}}.

\bibitem[Liu et~al.(2024{\natexlab{b}})Liu, Li, Li, Li, Zhang, Shen, and Lee]{liu2024llavanext}
Haotian Liu, Chunyuan Li, Yuheng Li, Bo Li, Yuanhan Zhang, Sheng Shen, and Yong~Jae Lee.
\newblock Llava-next: Improved reasoning, ocr, and world knowledge, 2024{\natexlab{b}}.

\bibitem[Liu et~al.(2024{\natexlab{c}})Liu, Yan, Zaharia, and Abbeel]{liu2024world}
Hao Liu, Wilson Yan, Matei Zaharia, and Pieter Abbeel.
\newblock World model on million-length video and language with ringattention.
\newblock \emph{arXiv preprint arXiv:2402.08268}, 2024{\natexlab{c}}.

\bibitem[Liu et~al.(2024{\natexlab{d}})Liu, Duan, Zhang, Li, Zhang, Zhao, Yuan, Wang, He, Liu, et~al.]{liu2023mmbench}
Yuan Liu, Haodong Duan, Yuanhan Zhang, Bo Li, Songyang Zhang, Wangbo Zhao, Yike Yuan, Jiaqi Wang, Conghui He, Ziwei Liu, et~al.
\newblock Mmbench: Is your multi-modal model an all-around player?
\newblock In \emph{ECCV}, 2024{\natexlab{d}}.

\bibitem[Loshchilov(2019)]{loshchilov2017decoupled}
I Loshchilov.
\newblock Decoupled weight decay regularization.
\newblock In \emph{ICLR}, 2019.

\bibitem[Lu et~al.(2022{\natexlab{a}})Lu, Clark, Zellers, Mottaghi, and Kembhavi]{lu2022unified}
Jiasen Lu, Christopher Clark, Rowan Zellers, Roozbeh Mottaghi, and Aniruddha Kembhavi.
\newblock Unified-io: A unified model for vision, language, and multi-modal tasks.
\newblock In \emph{ICLR}, 2022{\natexlab{a}}.

\bibitem[Lu et~al.(2024)Lu, Clark, Lee, Zhang, Khosla, Marten, Hoiem, and Kembhavi]{lu2024unified}
Jiasen Lu, Christopher Clark, Sangho Lee, Zichen Zhang, Savya Khosla, Ryan Marten, Derek Hoiem, and Aniruddha Kembhavi.
\newblock Unified-io 2: Scaling autoregressive multimodal models with vision language audio and action.
\newblock In \emph{CVPR}, 2024.

\bibitem[Lu et~al.(2022{\natexlab{b}})Lu, Mishra, Xia, Qiu, Chang, Zhu, Tafjord, Clark, and Kalyan]{lu2022learn}
Pan Lu, Swaroop Mishra, Tanglin Xia, Liang Qiu, Kai-Wei Chang, Song-Chun Zhu, Oyvind Tafjord, Peter Clark, and Ashwin Kalyan.
\newblock Learn to explain: Multimodal reasoning via thought chains for science question answering.
\newblock In \emph{NeurIPS}, 2022{\natexlab{b}}.

\bibitem[Masry et~al.(2022)Masry, Long, Tan, Joty, and Hoque]{masry2022chartqa}
Ahmed Masry, Do~Xuan Long, Jia~Qing Tan, Shafiq Joty, and Enamul Hoque.
\newblock Chartqa: A benchmark for question answering about charts with visual and logical reasoning.
\newblock In \emph{ACL}, 2022.

\bibitem[McKinzie et~al.(2024)McKinzie, Gan, Fauconnier, Dodge, Zhang, Dufter, Shah, Du, Peng, Weers, et~al.]{mckinzie2024mm1}
Brandon McKinzie, Zhe Gan, Jean-Philippe Fauconnier, Sam Dodge, Bowen Zhang, Philipp Dufter, Dhruti Shah, Xianzhi Du, Futang Peng, Floris Weers, et~al.
\newblock Mm1: Methods, analysis \& insights from multimodal llm pre-training.
\newblock \emph{arXiv preprint arXiv:2403.09611}, 2024.

\bibitem[Miech et~al.(2019)Miech, Zhukov, Alayrac, Tapaswi, Laptev, and Sivic]{miech2019howto100m}
Antoine Miech, Dimitri Zhukov, Jean-Baptiste Alayrac, Makarand Tapaswi, Ivan Laptev, and Josef Sivic.
\newblock Howto100m: Learning a text-video embedding by watching hundred million narrated video clips.
\newblock In \emph{ICCV}, 2019.

\bibitem[OpenAI(2024)]{OpenAI2024gpt4o}
OpenAI.
\newblock gpt4o, 2024.

\bibitem[Pan et~al.(2024{\natexlab{a}})Pan, Zhang, Tomlin, Zhou, Levine, and Suhr]{pan2024autonomous}
Jiayi Pan, Yichi Zhang, Nicholas Tomlin, Yifei Zhou, Sergey Levine, and Alane Suhr.
\newblock Autonomous evaluation and refinement of digital agents.
\newblock In \emph{COLM}, 2024{\natexlab{a}}.

\bibitem[Pan et~al.(2024{\natexlab{b}})Pan, Dong, Huang, Peng, Chen, and Wei]{pan2023kosmos}
Xichen Pan, Li Dong, Shaohan Huang, Zhiliang Peng, Wenhu Chen, and Furu Wei.
\newblock Kosmos-g: Generating images in context with multimodal large language models.
\newblock In \emph{ICLR}, 2024{\natexlab{b}}.

\bibitem[Preechakul et~al.(2022)Preechakul, Chatthee, Wizadwongsa, and Suwajanakorn]{preechakul2022diffusion}
Konpat Preechakul, Nattanat Chatthee, Suttisak Wizadwongsa, and Supasorn Suwajanakorn.
\newblock Diffusion autoencoders: Toward a meaningful and decodable representation.
\newblock In \emph{CVPR}, 2022.

\bibitem[Radford et~al.(2021)Radford, Kim, Hallacy, Ramesh, Goh, Agarwal, Sastry, Askell, Mishkin, Clark, et~al.]{radford2021learning}
Alec Radford, Jong~Wook Kim, Chris Hallacy, Aditya Ramesh, Gabriel Goh, Sandhini Agarwal, Girish Sastry, Amanda Askell, Pamela Mishkin, Jack Clark, et~al.
\newblock Learning transferable visual models from natural language supervision.
\newblock In \emph{ICML}, 2021.

\bibitem[Rajbhandari et~al.(2020)Rajbhandari, Rasley, Ruwase, and He]{rajbhandari2020zero}
Samyam Rajbhandari, Jeff Rasley, Olatunji Ruwase, and Yuxiong He.
\newblock Zero: Memory optimizations toward training trillion parameter models.
\newblock In \emph{SC20: International Conference for High Performance Computing, Networking, Storage and Analysis}, pages 1--16. IEEE, 2020.

\bibitem[Roberts et~al.(2019)Roberts, Raffel, Lee, Matena, Shazeer, Liu, Narang, Li, and Zhou]{roberts2019exploring}
Adam Roberts, Colin Raffel, Katherine Lee, Michael Matena, Noam Shazeer, Peter~J Liu, Sharan Narang, Wei Li, and Yanqi Zhou.
\newblock Exploring the limits of transfer learning with a unified text-to-text transformer.
\newblock \emph{JMLR}, 2019.

\bibitem[Rombach et~al.(2022)Rombach, Blattmann, Lorenz, Esser, and Ommer]{Rombach_2022_CVPR}
Robin Rombach, Andreas Blattmann, Dominik Lorenz, Patrick Esser, and Bj\"orn Ommer.
\newblock High-resolution image synthesis with latent diffusion models.
\newblock In \emph{CVPR}, 2022.

\bibitem[Schuhmann et~al.(2022)Schuhmann, Beaumont, Vencu, Gordon, Wightman, Cherti, Coombes, Katta, Mullis, Wortsman, et~al.]{schuhmann2022laion}
Christoph Schuhmann, Romain Beaumont, Richard Vencu, Cade Gordon, Ross Wightman, Mehdi Cherti, Theo Coombes, Aarush Katta, Clayton Mullis, Mitchell Wortsman, et~al.
\newblock Laion-5b: An open large-scale dataset for training next generation image-text models.
\newblock In \emph{NeurIPS}, 2022.

\bibitem[Shao et~al.(2024)Shao, Qian, Xiao, Song, Zong, Wang, Liu, and Li]{shao2024visual}
Hao Shao, Shengju Qian, Han Xiao, Guanglu Song, Zhuofan Zong, Letian Wang, Yu Liu, and Hongsheng Li.
\newblock Visual cot: Advancing multi-modal language models with a comprehensive dataset and benchmark for chain-of-thought reasoning.
\newblock In \emph{NeurIPS}, 2024.

\bibitem[Sidorov et~al.(2020)Sidorov, Hu, Rohrbach, and Singh]{sidorov2020textcaps}
Oleksii Sidorov, Ronghang Hu, Marcus Rohrbach, and Amanpreet Singh.
\newblock Textcaps: a dataset for image captioning with reading comprehension, 2020.

\bibitem[Sun et~al.(2024{\natexlab{a}})Sun, Cui, Zhang, Zhang, Yu, Wang, Rao, Liu, Huang, and Wang]{sun2024generative}
Quan Sun, Yufeng Cui, Xiaosong Zhang, Fan Zhang, Qiying Yu, Yueze Wang, Yongming Rao, Jingjing Liu, Tiejun Huang, and Xinlong Wang.
\newblock Generative multimodal models are in-context learners.
\newblock In \emph{CVPR}, 2024{\natexlab{a}}.

\bibitem[Sun et~al.(2024{\natexlab{b}})Sun, Yu, Cui, Zhang, Zhang, Wang, Gao, Liu, Huang, and Wang]{sun2023generative}
Quan Sun, Qiying Yu, Yufeng Cui, Fan Zhang, Xiaosong Zhang, Yueze Wang, Hongcheng Gao, Jingjing Liu, Tiejun Huang, and Xinlong Wang.
\newblock Generative pretraining in multimodality.
\newblock In \emph{ICLR}, 2024{\natexlab{b}}.

\bibitem[Taori et~al.(2023)Taori, Gulrajani, Zhang, Dubois, Li, Guestrin, Liang, and Hashimoto]{Taori2023Alpaca}
Rohan Taori, Ishaan Gulrajani, Tianyi Zhang, Yann Dubois, Xuechen Li, Carlos Guestrin, Percy Liang, and Tatsunori~B. Hashimoto.
\newblock Alpaca: A strong, replicable instruction-following model, 2023.

\bibitem[Team(2024)]{team2024chameleon}
Chameleon Team.
\newblock Chameleon: Mixed-modal early-fusion foundation models.
\newblock \emph{arXiv preprint arXiv:2405.09818}, 2024.

\bibitem[Tong et~al.(2024{\natexlab{a}})Tong, Brown, Wu, Woo, Middepogu, Akula, Yang, Yang, Iyer, Pan, et~al.]{tong2024cambrian}
Shengbang Tong, Ellis Brown, Penghao Wu, Sanghyun Woo, Manoj Middepogu, Sai~Charitha Akula, Jihan Yang, Shusheng Yang, Adithya Iyer, Xichen Pan, et~al.
\newblock Cambrian-1: A fully open, vision-centric exploration of multimodal llms.
\newblock In \emph{NeurIPS}, 2024{\natexlab{a}}.

\bibitem[Tong et~al.(2024{\natexlab{b}})Tong, Jones, and Steinhardt]{tong2024mass}
Shengbang Tong, Erik Jones, and Jacob Steinhardt.
\newblock Mass-producing failures of multimodal systems with language models.
\newblock In \emph{NeurIPS}, 2024{\natexlab{b}}.

\bibitem[Tong et~al.(2024{\natexlab{c}})Tong, Liu, Zhai, Ma, LeCun, and Xie]{tong2024eyes}
Shengbang Tong, Zhuang Liu, Yuexiang Zhai, Yi Ma, Yann LeCun, and Saining Xie.
\newblock Eyes wide shut? exploring the visual shortcomings of multimodal llms.
\newblock In \emph{CVPR}, 2024{\natexlab{c}}.

\bibitem[Touvron et~al.(2023)Touvron, Martin, Stone, Albert, Almahairi, Babaei, Bashlykov, Batra, Bhargava, Bhosale, et~al.]{touvron2023llama2}
Hugo Touvron, Louis Martin, Kevin Stone, Peter Albert, Amjad Almahairi, Yasmine Babaei, Nikolay Bashlykov, Soumya Batra, Prajjwal Bhargava, Shruti Bhosale, et~al.
\newblock {LLaMA} 2: Open foundation and fine-tuned chat models.
\newblock 2023.

\bibitem[Wang et~al.(2024{\natexlab{a}})Wang, Bai, Tan, Wang, Fan, Bai, Chen, Liu, Wang, Ge, et~al.]{wang2024qwen2}
Peng Wang, Shuai Bai, Sinan Tan, Shijie Wang, Zhihao Fan, Jinze Bai, Keqin Chen, Xuejing Liu, Jialin Wang, Wenbin Ge, et~al.
\newblock Qwen2-vl: Enhancing vision-language model's perception of the world at any resolution.
\newblock \emph{arXiv preprint arXiv:2409.12191}, 2024{\natexlab{a}}.

\bibitem[Wang et~al.(2024{\natexlab{b}})Wang, Zhang, Luo, Sun, Cui, Wang, Zhang, Wang, Li, Yu, et~al.]{wang2024emu3}
Xinlong Wang, Xiaosong Zhang, Zhengxiong Luo, Quan Sun, Yufeng Cui, Jinsheng Wang, Fan Zhang, Yueze Wang, Zhen Li, Qiying Yu, et~al.
\newblock Emu3: Next-token prediction is all you need.
\newblock \emph{arXiv preprint arXiv:2409.18869}, 2024{\natexlab{b}}.

\bibitem[Wei et~al.(2022{\natexlab{a}})Wei, Bosma, Zhao, Guu, Yu, Lester, Du, Dai, and Le]{wei2021finetuned}
Jason Wei, Maarten Bosma, Vincent~Y Zhao, Kelvin Guu, Adams~Wei Yu, Brian Lester, Nan Du, Andrew~M Dai, and Quoc~V Le.
\newblock Finetuned language models are zero-shot learners.
\newblock In \emph{ICLR}, 2022{\natexlab{a}}.

\bibitem[Wei et~al.(2022{\natexlab{b}})Wei, Wang, Schuurmans, Bosma, Xia, Chi, Le, Zhou, et~al.]{wei2022chain}
Jason Wei, Xuezhi Wang, Dale Schuurmans, Maarten Bosma, Fei Xia, Ed Chi, Quoc~V Le, Denny Zhou, et~al.
\newblock Chain-of-thought prompting elicits reasoning in large language models.
\newblock In \emph{NeurIPS}, 2022{\natexlab{b}}.

\bibitem[Wu et~al.(2024{\natexlab{a}})Wu, Chen, Wu, Ma, Liu, Pan, Liu, Xie, Yu, Ruan, et~al.]{wu2024janus}
Chengyue Wu, Xiaokang Chen, Zhiyu Wu, Yiyang Ma, Xingchao Liu, Zizheng Pan, Wen Liu, Zhenda Xie, Xingkai Yu, Chong Ruan, et~al.
\newblock Janus: Decoupling visual encoding for unified multimodal understanding and generation.
\newblock \emph{arXiv preprint arXiv:2410.13848}, 2024{\natexlab{a}}.

\bibitem[Wu and Xie(2024)]{wu2023vstar}
Penghao Wu and Saining Xie.
\newblock V*: Guided visual search as a core mechanism in multimodal llms.
\newblock In \emph{CVPR}, 2024.

\bibitem[Wu et~al.(2024{\natexlab{b}})Wu, Zhang, Chen, Tang, Li, Fang, Zhu, Xie, Yin, Yi, et~al.]{wu2024vila}
Yecheng Wu, Zhuoyang Zhang, Junyu Chen, Haotian Tang, Dacheng Li, Yunhao Fang, Ligeng Zhu, Enze Xie, Hongxu Yin, Li Yi, et~al.
\newblock Vila-u: a unified foundation model integrating visual understanding and generation.
\newblock \emph{arXiv preprint arXiv:2409.04429}, 2024{\natexlab{b}}.

\bibitem[xAI(2024)]{grok}
xAI.
\newblock grok, 2024.

\bibitem[Xie et~al.(2024)Xie, Mao, Bai, Zhang, Wang, Lin, Gu, Chen, Yang, and Shou]{xie2024show}
Jinheng Xie, Weijia Mao, Zechen Bai, David~Junhao Zhang, Weihao Wang, Kevin~Qinghong Lin, Yuchao Gu, Zhijie Chen, Zhenheng Yang, and Mike~Zheng Shou.
\newblock Show-o: One single transformer to unify multimodal understanding and generation.
\newblock \emph{arXiv preprint arXiv:2408.12528}, 2024.

\bibitem[Xu et~al.(2024)Xu, Xie, Tan, Huang, Howes, Sharma, Li, Ghosh, Zettlemoyer, and Feichtenhofer]{xu2023demystifying}
Hu Xu, Saining Xie, Xiaoqing~Ellen Tan, Po-Yao Huang, Russell Howes, Vasu Sharma, Shang-Wen Li, Gargi Ghosh, Luke Zettlemoyer, and Christoph Feichtenhofer.
\newblock Demystifying clip data.
\newblock In \emph{ICLR}, 2024.

\bibitem[Ye et~al.(2024)Ye, Xu, Li, and {Allen-Zhu}]{YXLA2024-gsm1}
Tian Ye, Zicheng Xu, Yuanzhi Li, and Zeyuan {Allen-Zhu}.
\newblock {Physics of Language Models: Part 2.1, Grade-School Math and the Hidden Reasoning Process}.
\newblock \emph{ArXiv e-prints}, abs/2407.20311, 2024.
\newblock Full version available at \url{http://arxiv.org/abs/2407.20311}.

\bibitem[Yue et~al.(2024{\natexlab{a}})Yue, Ni, Zhang, Zheng, Liu, Zhang, Stevens, Jiang, Ren, Sun, et~al.]{yue2023mmmu}
Xiang Yue, Yuansheng Ni, Kai Zhang, Tianyu Zheng, Ruoqi Liu, Ge Zhang, Samuel Stevens, Dongfu Jiang, Weiming Ren, Yuxuan Sun, et~al.
\newblock Mmmu: A massive multi-discipline multimodal understanding and reasoning benchmark for expert agi.
\newblock In \emph{CVPR}, 2024{\natexlab{a}}.

\bibitem[Yue et~al.(2024{\natexlab{b}})Yue, Zheng, Ni, Wang, Zhang, Tong, Sun, Yin, Yu, Zhang, et~al.]{yue2024mmmu}
Xiang Yue, Tianyu Zheng, Yuansheng Ni, Yubo Wang, Kai Zhang, Shengbang Tong, Yuxuan Sun, Ming Yin, Botao Yu, Ge Zhang, et~al.
\newblock Mmmu-pro: A more robust multi-discipline multimodal understanding benchmark.
\newblock \emph{arXiv preprint arXiv:2409.02813}, 2024{\natexlab{b}}.

\bibitem[Yuksekgonul et~al.(2022)Yuksekgonul, Bianchi, Kalluri, Jurafsky, and Zou]{yuksekgonul2022and}
Mert Yuksekgonul, Federico Bianchi, Pratyusha Kalluri, Dan Jurafsky, and James Zou.
\newblock When and why vision-language models behave like bags-of-words, and what to do about it?
\newblock In \emph{ICLR}, 2022.

\bibitem[Zhai et~al.(2023)Zhai, Mustafa, Kolesnikov, and Beyer]{zhai2023sigmoid}
Xiaohua Zhai, Basil Mustafa, Alexander Kolesnikov, and Lucas Beyer.
\newblock Sigmoid loss for language image pre-training.
\newblock In \emph{ICCV}, 2023.

\bibitem[Zhai et~al.(2024)Zhai, Bai, Lin, Pan, Tong, Zhou, Suhr, Xie, LeCun, Ma, et~al.]{zhai2024fine}
Yuexiang Zhai, Hao Bai, Zipeng Lin, Jiayi Pan, Shengbang Tong, Yifei Zhou, Alane Suhr, Saining Xie, Yann LeCun, Yi Ma, et~al.
\newblock Fine-tuning large vision-language models as decision-making agents via reinforcement learning.
\newblock In \emph{NeurIPS}, 2024.

\bibitem[Zhang et~al.(2024)Zhang, Gui, Sun, Feng, Xu, Zhang, Fu, Li, Hauptmann, Bisk, et~al.]{zhang2024direct}
Ruohong Zhang, Liangke Gui, Zhiqing Sun, Yihao Feng, Keyang Xu, Yuanhan Zhang, Di Fu, Chunyuan Li, Alexander Hauptmann, Yonatan Bisk, et~al.
\newblock Direct preference optimization of video large multimodal models from language model reward.
\newblock \emph{arXiv preprint arXiv:2404.01258}, 2024.

\bibitem[Zhang et~al.(2023)Zhang, McKinzie, Gan, Shankar, and Toshev]{zhang2023pre}
Yuhui Zhang, Brandon McKinzie, Zhe Gan, Vaishaal Shankar, and Alexander Toshev.
\newblock Pre-trained language models do not help auto-regressive text-to-image generation.
\newblock In \emph{EMNLP}, 2023.

\bibitem[Zhou et~al.(2024{\natexlab{a}})Zhou, Liu, Xu, Iyer, Sun, Mao, Ma, Efrat, Yu, Yu, et~al.]{zhou2024lima}
Chunting Zhou, Pengfei Liu, Puxin Xu, Srinivasan Iyer, Jiao Sun, Yuning Mao, Xuezhe Ma, Avia Efrat, Ping Yu, Lili Yu, et~al.
\newblock Lima: Less is more for alignment.
\newblock In \emph{NeurIPS}, 2024{\natexlab{a}}.

\bibitem[Zhou et~al.(2024{\natexlab{b}})Zhou, Yu, Babu, Tirumala, Yasunaga, Shamis, Kahn, Ma, Zettlemoyer, and Levy]{zhou2024transfusion}
Chunting Zhou, Lili Yu, Arun Babu, Kushal Tirumala, Michihiro Yasunaga, Leonid Shamis, Jacob Kahn, Xuezhe Ma, Luke Zettlemoyer, and Omer Levy.
\newblock Transfusion: Predict the next token and diffuse images with one multi-modal model.
\newblock \emph{arXiv preprint arXiv:2408.11039}, 2024{\natexlab{b}}.

\bibitem[Zohar et~al.(2024)Zohar, Wang, Bitton, Szpektor, and Yeung-levy]{zohar2024videostar}
Orr Zohar, Xiaohan Wang, Yonatan Bitton, Idan Szpektor, and Serena Yeung-levy.
\newblock Video-star: Self-training enables video instruction tuning with any supervision.
\newblock In \emph{arXiv preprint arXiv:2407.06189}, 2024.

\end{thebibliography}
}
\clearpage
\newpage
\beginappendix

\section{Training Details and Hyperparameters} \label{appendix: training hyperparameters}

\subsection{\ours{} Training} \label{appendix: MetaMorph Training}

We follow the training recipe outlined in prior studies~\citep{tong2024cambrian, mckinzie2024mm1}, using a two-stage training approach. First, we pretrain a two-layer MLP with a GELU activation~\citep{hendrycks2016gaussian} as the adapter between the visual tokens and the LLM. We train this adapter on Cambrian adapter data while excluding all data points sourced from LAION~\citep{schuhmann2022laion}. Next, we finetune the entire model, excluding the vision backbone, using the instruction tuning data described in \cref{sec: data} and detailed in \cref{appendix: Data}.

We use DeepSpeed~\citep{rajbhandari2020zero} Zero-3 to train our model on H100 GPUs. Detailed training hyperparameters for all experiments are provided in \cref{tab: metamorph training}. We conduct all of the experiments with 1 epoch.

\begin{table*}[h]
\centering
    \small  
    \setlength\tabcolsep{2pt} 
    \begin{adjustbox}{max width=\textwidth}
    \begin{tabular}{l|l|cc|ccc|ccc}
     \multicolumn{1}{c|}{} & \multicolumn{1}{c|}{Backbone} & \multicolumn{2}{c|}{Data} & \multicolumn{3}{c|}{Adapter} &  \multicolumn{3}{c}{Instruction Tuning}\\
      Experiment  & \multicolumn{1}{c|}{LLM} & Adapter & Instruction Tuning &     lr & wd & bs & lr & wd & bs \\
      \hline
      \cref{sec: teach LLM to generate vision} (LLaMA-3 8B) &  LLaMA-3 8B & Cambrian Adapter Data$^*$ & \cref{sec: teach LLM to generate vision} Experiment Setting & 4.90e-5 & 0.0 & 768 & 6.93e-5 & 0 & 1536 \\
      \cref{sec: teach LLM to generate vision} (LLaMA-3.1 8B) &  LLaMA-3.1 8B & Cambrian Adapter Data$^*$ & \cref{sec: teach LLM to generate vision} Experiment Setting & 4.90e-5 & 0.0 & 768 & 6.93e-5 & 0 & 1536 \\
      \cref{sec: teach LLM to generate vision} (LLaMA-3 70B) &  LLaMA-3 70B & Cambrian Adapter Data$^*$ & \cref{sec: teach LLM to generate vision} Experiment Setting & 4.90e-5 & 0.0 & 768 & 4.90e-5 & 0 & 768 \\
       \ours{} & LLaMA-3.1 8B & Cambrian Adapter Data$^*$ & All Data from \cref{sec: data} & 4.90e-5 & 0.0 & 768 & 6.93e-5 & 0 & 1536 \\
    \end{tabular}
    \end{adjustbox}
         \vspace{-1em}
\caption{\small \textbf{Implementation details and hyperparameters for all experiments.} $^*$We exclude data points in LAION~\citep{schuhmann2022laion} from Cambrian adapter data. 
}
\label{tab: metamorph training}
\end{table*}

\subsection{Diffusion Visualizer Training} \label{appendix: Diffusion Training}
We leverage pretrained diffusion models such as Stable Diffusion 1.5~\citep{Rombach_2022_CVPR}. We use a 2-layer MLP projector to align the SigLIP embedding dimension with the cross-attention dimension in the pretrained diffusion model. The first layer applies a linear transformation to map the input dimension to 2048, followed by layer normalization~\citep{ba2016layer} and a ReLU activation. The second layer reduces the 2048-dimensional features to the output dimension through a linear transformation, followed by a final layernorm.

We set the batch size to 2112. The learning rate schedule begins with a logarithmic warm-up over the first 2000 steps, gradually increasing from zero to a peak value of 1.1e-5. After this warm-up phase, the learning rate decreases linearly over the next 12000 steps until reaching zero. We use the AdamW~\citep{loshchilov2017decoupled} optimizer to train our model, with \(\beta\) parameters \((0.9, 0.999)\). We apply a weight decay of 0.01. 

During diffusion training, we freeze the VAE encoder and Siglip encoder, only training the projector and the diffusion U-Net. The CFG level is set to 0.7. This is because we start with a pretrained diffusion model and aim to transform the conditioning from CLIP text to SigLIP image embeddings. A higher CFG level ensures the model maintains high image quality while gradually adapting to the new conditioning in the remaining fraction. Empirically, this approach achieves the best balance between adaptation and image quality. For the training datasets, since we finetune the diffusion model to condition on SigLIP image embeddings, training this model does not require text descriptions for conditioning. Instead, we use images curated through in MetaCLIP~\citep{xu2023demystifying} and train this diffusion model to visualize the visual tokens generated by \ours{}.

\subsection{Evaluation Benchmarks}
For evaluation, we use nine ImageQA, one VideoQA and two generation benchmarks:

\begin{itemize}
    \item \textbf{MMBench}~\citep{liu2023mmbench}: A comprehensive benchmark spans across 20 multimodal ability dimensions.
    \item \textbf{Seed}~\citep{ge2023planting}: A benchmark focusing on visual tasks for multimodal understanding, consists of 19k multiple choice questions with accurate human annotations.
    \item \textbf{V*STAR}~\citep{wu2023vstar}: A VQA benchmark designed for testing details in high-resolution images.
    \item \textbf{MMVP}~\citep{tong2024eyes}: A benchmark for evaluating ``CLIP-Blind'' pairs in Vision Language Models.
    \item \textbf{MMMU}~\citep{yue2023mmmu}: A benchmark designed to evaluate multimodal models on extensive multi-discipline tasks requiring college-level subject knowledge and deliberate reasoning.
    \item \textbf{ChartQA}~\citep{masry2022chartqa}: A large-scale benchmark involving visual and logical reasoning over charts.
    \item \textbf{TextVQA}~\citep{sidorov2020textcaps}:A benchmark designed to evaluate models' ability to read and reason about text in images to answer questions.
    \item \textbf{ScienceQA}~\citep{lu2022learn}: A multimodal benchmark for answering science-related questions requiring integration of visual and textual data.
    \item \textbf{RealWorldQA}~\citep{grok}: A benchmark focused on real-world multimodal reasoning tasks.
    \item \textbf{MV-Bench}~\citep{li2024mvbench}: A benchmark contains a comprehensive video understanding benchmark, which covers 20 challenging video tasks that cannot be effectively solved with a single frame.
    \item \textbf{FID Score}~\citep{heusel2017gans}: A metric for evaluating the quality of generated images by comparing their feature distributions with real images.
    \item \textbf{CLIP Score}~\citep{hessel2021clipscore}: A benchmark metric that uses CLIP embeddings to measure alignment between generated images and their corresponding text descriptions.
\end{itemize}

\section{Ablation Studies on Visual Prediction Objective} \label{appendix: vision prediction ablation}
We compare our approach to the commonly used L1 regression loss, which has been widely adopted in contrastive self-supervised learning methods~\citep{lecun2022path, bardes2024revisiting}. For this comparison, we train \ours{}, based on LLaMA-3 8B, using datasets described in \cref{sec: data}. We highlight that cosine similarity and L1 loss influence the embedding outputs differently: cosine similarity enforces normalization, while L1 loss does not. This discrepancy in output normalization prevents a direct and fair comparison in terms of generation performance. Consequently, our analysis focuses exclusively on VQA performance.

In \cref{tab: ablate loss}, we compare models trained using L1 loss and cosine similarity loss. Our analysis reveals that training with cosine similarity results in better average performance and outperforms L1 loss on most benchmarks. Notably, these vision loss functions affect only tasks requiring visual predictions and do not directly influence VQA tasks, as the VQA training data does not include image token responses. This improvement is potentially because training with cosine similarity enhances visual generation, which in turn contributes to better visual understanding. 

To further investigate, we compare our method—incorporating a broader range of non-VQA data alongside Cambrian-7M—--with a baseline trained exclusively on Cambrian-7M. The results show that combining broader dataset with cosine similarity loss leads to better performance across multiple benchmarks. This finding reinforces our earlier observations in \cref{sec: teach LLM to generate vision}: enhancing visual generation capabilities contributes to improved visual understanding, highlighting the benefits of leveraging non-VQA data.

\begin{table*}[t]
    \centering
    \small 
    \begin{tabular}{r|c|ccccccccc}
     \multicolumn{1}{c}{Loss} &
     \multicolumn{10}{c}{Image QA} \\
      &
      \rotatebox{90}{AVG} &
      \rotatebox{90}{MMBench$^\text{EN}$} &
      \rotatebox{90}{SEED} &
      \rotatebox{90}{RealworldQA} &
      \rotatebox{90}{MMVP} &
      \rotatebox{90}{SQA} &
      \rotatebox{90}{MMMU} &
      \rotatebox{90}{VStar} &
      \rotatebox{90}{ChartQA} &
      \rotatebox{90}{TextVQA} \\
           \hline
     \color{gray} None (VQA Only) &  \color{gray}55.50 &  \color{gray}73.11 & \color{gray} 69.96 &  \color{gray}55.69 &  \color{gray}41.33 &  \color{gray}80.39 & \color{gray} 37.29 &  \color{gray}46.60 &  \color{gray}35.16 &  \color{gray}59.96\\ 
           
     L1 Loss & 53.83 & 72.17 & 69.28 & 57.25 & 34.67 & 79.00 & 34.00 & 45.55 & 32.40 & 60.17\\ 
     Cosine Sim & 55.93 & 73.78 & 71.36 & 55.03 & 44.00 & 79.83 & 35.29 & 47.64 & 36.60 & 59.79    \\

    \end{tabular}.
     \vspace{-1em}
 \caption{\small \textbf{Comparison of different loss functions.} Training with cosine similarity loss enables the model to effectively utilize non-VQA data, which in turn enhances its visual understanding. }
\label{tab: ablate loss}
\end{table*}

\section{Data} \label{appendix: Data}

\subsection{Data Composition}
\begin{figure}[t]
    \centering
    \begin{minipage}{0.27\columnwidth}
        \includegraphics[width=\columnwidth]{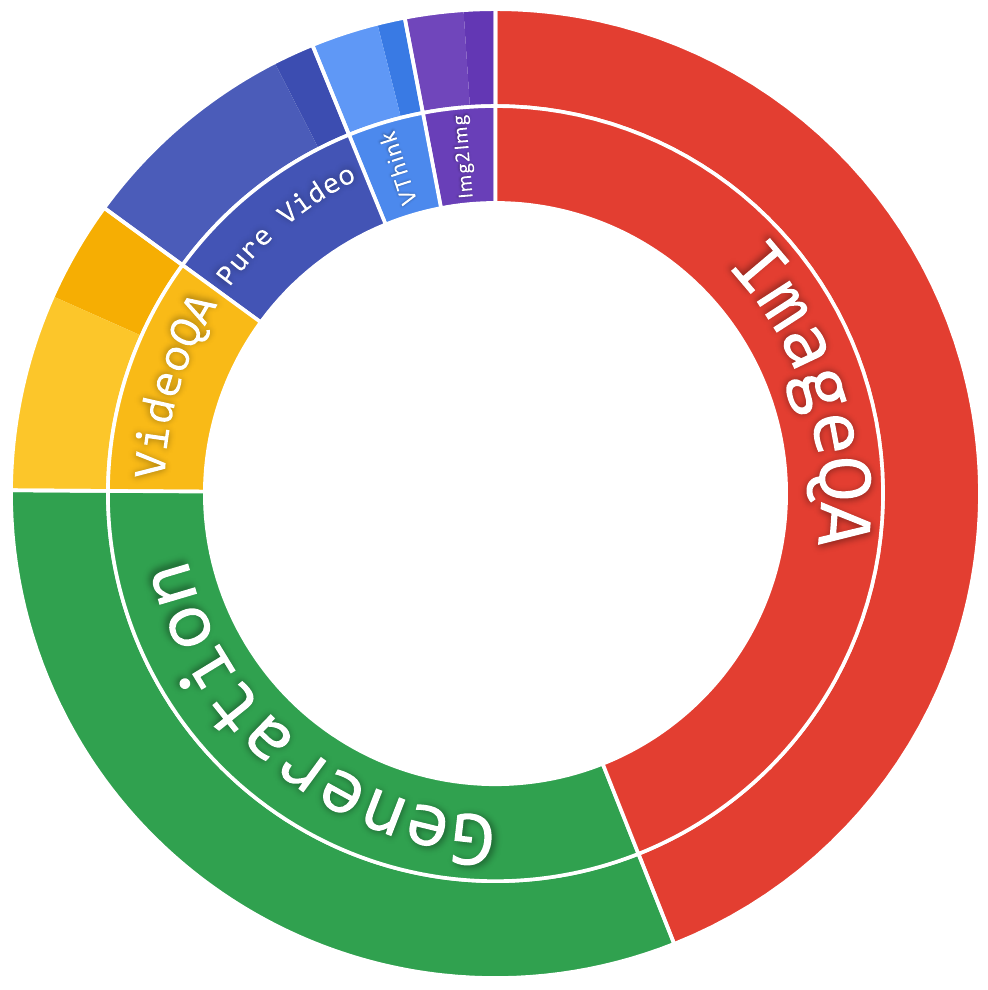}
    \end{minipage}%
    \hfill
    \begin{minipage}{0.70\columnwidth}
        \centering
        \renewcommand{\arraystretch}{1.1}
        \setlength\tabcolsep{2pt}
        \fontsize{5.8pt}{10pt}\selectfont
        \begin{tabular}{@{}p{0.5\columnwidth}p{0.5\columnwidth}@{}}
        \makecell{\cellcolor[RGB]{227,62,49} \textcolor{white}{ImageQA (44.0\%)}} &
        \makecell{\cellcolor[RGB]{48,161,79} \textcolor{white}{Generation (31.1\%)}} \\
        
        \tikz[baseline=0.05em] \fill [color={rgb,255: red,227; green,61; blue,49}] (0,0) rectangle (0.75em,0.75em); Cambrian-7M~\citep{tong2024cambrian} (7067.0 K) &
        \tikz[baseline=0.05em] \fill [color={rgb,255: red,48; green,161; blue,79}] (0,0) rectangle (0.75em,0.75em); MetaCLIP~\citep{xu2023demystifying} (5000.0 K) \\
        
        \makecell{\cellcolor[RGB]{249,186,23} \textcolor{white}{VideoQA (9.9\%)}} &
        \makecell{\cellcolor[RGB]{76,137,237} \textcolor{white}{Visual Thinking (3.2\%)}} \\
        
        \tikz[baseline=0.05em] \fill [color={rgb,255: red,251; green,198; blue,42}] (0,0) rectangle (0.75em,0.75em); VideoStar~\citep{zohar2024videostar} (1055.0 K) &
        \tikz[baseline=0.05em] \fill [color={rgb,255: red,94; green,151; blue,245}] (0,0) rectangle (0.75em,0.75em); VisualCoT~\citep{shao2024visual} (361.0 K) \\
        
        \tikz[baseline=0.05em] \fill [color={rgb,255: red,246; green,173; blue,3}] (0,0) rectangle (0.75em,0.75em); ShareVideo~\citep{zhang2024direct} (540.0 K) &
        \tikz[baseline=0.05em] \fill [color={rgb,255: red,57; green,122; blue,228}] (0,0) rectangle (0.75em,0.75em); VStar~\citep{wu2023vstar} (148.0 K) \\
        
        \makecell{\cellcolor[RGB]{67,84,181} \textcolor{white}{Pure Video (8.8\%)}} &
        \makecell{\cellcolor[RGB]{105,63,184} \textcolor{white}{Image-to-Image (3.0\%)}} \\
        
        \tikz[baseline=0.05em] \fill [color={rgb,255: red,74; green,91; blue,185}] (0,0) rectangle (0.75em,0.75em); HowTo100M~\citep{miech2019howto100m} (1193.0 K) &
        \tikz[baseline=0.05em] \fill [color={rgb,255: red,112; green,69; blue,187}] (0,0) rectangle (0.75em,0.75em); InstructPix2Pix~\citep{brooks2023instructpix2pix} (313.0 K) \\
        \tikz[baseline=0.05em] \fill [color={rgb,255: red,60; green,77; blue,176}] (0,0) rectangle (0.75em,0.75em); SmthSmthV2~\citep{goyal2017something} (220.0 K) &
        \tikz[baseline=0.05em] \fill [color={rgb,255: red,98; green,55; blue,180}] (0,0) rectangle (0.75em,0.75em); Aurora~\citep{krojer2024learning} (169.0 K) \\
        \end{tabular}
    \end{minipage}
    \caption{\small \textbf{Data composition.} \textbf{Left:} The inner circle shows the distribution of \ours{} data. \textbf{Right:} All the data sources and categories in the \ours{} data.}
    \label{fig:datacomposition}
\end{figure}

We summarize the categorization of data and the number of samples for each source in \cref{fig:datacomposition}. This diverse dataset is curated to showcase that an LLM can be finetuned across a variety of tasks, where each task contributes to and enhances the performance of others, as discussed in \cref{sec: generation is triggered}.

\subsection{Data Proprocessing} \label{appendix: data preprocessing}
As discussed in \cref{sec: data}, we use a wide range of data, spanning from visual question answering tasks to unlabeled video data. Here, we detail the preprocessing steps applied to each data source to convert them into instruction-tuning-style QA conversations.

\paragraph{ImageQA.} We use Cambrian-7M~\citep{tong2024cambrian}, a dataset already curated in instruction tuning format. An example entry looks like the below:

\begin{tcolorbox}[
  colback=white, 
  colframe=black!50, 
  boxrule=0.3mm, 
  rounded corners, 
  title=Example from ImageQA,
  fonttitle=\bfseries, 
  left=2mm, right=2mm, top=2mm, bottom=2mm 
]
\textbf{Prompt: }

\texttt{\small <image\_start><image><image\_end>} What is the animal in the image? 

\textbf{Response:} 

It is a burmilla cat.
\end{tcolorbox}

\paragraph{VideoQA.} We use VideoStar~\citep{zohar2024videostar} and ShareVideo~\citep{chen2024sharegpt4video}, both curated in an instruction tuning format. For each video, we extract frames at a rate of one frame per second and input these frames into the LLM. An example QA entry for an 8-second video is structured as follows:
\begin{tcolorbox}[
  colback=white, 
  colframe=black!50, 
  boxrule=0.3mm, 
  rounded corners, 
  title=Example from VideoQA,
  fonttitle=\bfseries, 
  left=2mm, right=2mm, top=2mm, bottom=2mm 
]
\textbf{Prompt: }

\texttt{\small<image\_start><image><image\_end><image\_start><image><image\_end><image\_start><image><image\_end><image\_start><image><image\_end><image\_start><image><image\_end><image\_start><image><image\_end><image\_start><image><image\_end><image\_start><image><image\_end>}  

What's the color of the dog in this video? 
(a) white
(b) yellow
(c) black
Please only answer a single letter and nothing else

\textbf{Response:} 

b
\end{tcolorbox}

\paragraph{Generation data.}  We use image-text pairs in MetaCLIP~\citep{xu2023demystifying}. The original data consists of images paired with corresponding text descriptions. We add system prompts and define answering formats, transforming the image-text pairs into question-answer formats suitable for instruction tuning.

\begin{tcolorbox}[
  colback=white, 
  colframe=black!50, 
  boxrule=0.3mm, 
  rounded corners, 
  title=Example from Generation data,
  fonttitle=\bfseries, 
  left=2mm, right=2mm, top=2mm, bottom=2mm 
]
\textbf{Prompt: } 

Generate an image of a puppy. 

\textbf{Response:} 

Here is an image based on your request: \texttt{\small <image\_start><image><image\_end>}.
\end{tcolorbox}

Unlike in ImageQA and VideoQA, we require the model to predict the visual tokens in the response. 

\paragraph{Visual Thinking Data.} We explore incorporating vision as part of the model's reasoning process to enhance its answers. As a preliminary step, we experiment with the Visualization-of-Thought~\citep{shao2024visual} and VStar~\citep{wu2023vstar} datasets. Originally, these datasets were designed to teach models how to utilize external tools and APIs, such as segmentation or zoom-in cropping. In this work, we aim to integrate these visual skills directly into the model's inference steps. We use system prompts, such as \textit{``think visually before you answer the question''}, to activate this visual reasoning mode. Here is an example:
\begin{tcolorbox}[
  colback=white, 
  colframe=black!50, 
  boxrule=0.3mm, 
  rounded corners, 
  title=Example from Visual thinking data,
  fonttitle=\bfseries, 
  left=2mm, right=2mm, top=2mm, bottom=2mm 
]
\textbf{Prompt: } 

Think visually before you answer the question: what is the color of the hat in the image \texttt{\small <image\_start><image><image\_end>}. 

\textbf{Response:} 

Let's consider this visually: 
\texttt{\small <image\_start><image><image\_end>} black.
\end{tcolorbox}

In this example, the model first predicts a zoomed-in view of the key region in the image necessary for answering the question, and then it provides the answer.

\paragraph{Image to image data.}  The model is required to predict a transformed image based on a given text description. We use Aurora~\citep{krojer2024learning} and InstructPix2Pix~\citep{brooks2023instructpix2pix} datasets to train this capability. An example is presented below:

\begin{tcolorbox}[
  colback=white, 
  colframe=black!50, 
  boxrule=0.3mm, 
  rounded corners, 
  title=Example from Image to image data,
  fonttitle=\bfseries, 
  left=2mm, right=2mm, top=2mm, bottom=2mm 
]
\textbf{Prompt: } 

\texttt{\small <image\_start><image><image\_end>} Make this a rainy day. 

\textbf{Response:} 

\texttt{\small <image\_start><image><image\_end>}.
\end{tcolorbox}

Here, the model is expected to predict the visual tokens of the transformed image as per the specified prompt. 

\paragraph{Pure video data.} We explore commonly used open-source video datasets in instruction tuning: SomethingSomethingV2~\citep{goyal2017something} and HowTo100M~\citep{miech2019howto100m}. We design the following tasks from the pure video:

1) Forward Frame Prediction. In this task, the model is presented with the initial frame of a video sequence and must predict the subsequent frames at fixed time intervals. An example is presented below:
\begin{tcolorbox}[
  colback=white,
  colframe=black!50,
  boxrule=0.3mm,
  rounded corners,
  title=Example of Forward Frame Prediction,
  fonttitle=\bfseries,
  left=2mm, right=2mm, top=2mm, bottom=2mm
]
\textbf{Prompt: } 

\texttt{\small <image\_start><image><image\_end>} Can you predict what happens in the next 3 frames, each 5 seconds apart?

\textbf{Response:} 

\texttt{\small <image\_start><image><image\_end><image\_start><image><image\_end><image\_start><image><image\_end>}
\end{tcolorbox}

2) Partial Sequence Completion. This task requires the model to complete a video sequence when given only a subset of frames while maintaining temporal coherence:
\begin{tcolorbox}[
  colback=white,
  colframe=black!50,
  boxrule=0.3mm,
  rounded corners,
  title=Example of Partial Sequence Completion,
  fonttitle=\bfseries,
  left=2mm, right=2mm, top=2mm, bottom=2mm
]
\textbf{Prompt: } 

\texttt{\small <image\_start><image><image\_end><image\_start><image><image\_end><image\_start><image><image\_end>} 
Can you predict the 2 missing frames in this 5-second-interval sequence?

\textbf{Response:} 

\texttt{\small <image\_start><image><image\_end><image\_start><image><image\_end>}
\end{tcolorbox}

3) Reverse Temporal Prediction. This task challenges the model to reconstruct the preceding frames given the final frame of a sequence:
\begin{tcolorbox}[
  colback=white,
  colframe=black!50,
  boxrule=0.3mm,
  rounded corners,
  title=Example of Reverse Temporal Reasoning,
  fonttitle=\bfseries,
  left=2mm, right=2mm, top=2mm, bottom=2mm
]
\textbf{Prompt: } 

\texttt{\small <image\_start><image><image\_end>} Work backwards to predict the previous 4 frames, each 5 seconds apart.

\textbf{Response:} 

\texttt{\small <image\_start><image><image\_end><image\_start><image><image\_end><image\_start><image><image\_end><image\_start><image><image\_end>}
\end{tcolorbox}

4) Temporal Sequence Reordering. In this task, the model receives a shuffled sequence of video frames and must reconstruct their correct temporal order:
\begin{tcolorbox}[
  colback=white,
  colframe=black!50,
  boxrule=0.3mm,
  rounded corners,
  title=Example of Temporal Sequence Reordering,
  fonttitle=\bfseries,
  left=2mm, right=2mm, top=2mm, bottom=2mm
]
\textbf{Prompt: } 

\texttt{\small <image\_start><image><image\_end><image\_start><image><image\_end><image\_start><image><image\_end><image\_start><image><image\_end>} 

Arrange these frames in their correct temporal sequence.

\textbf{Response:} 

\texttt{\small <image\_start><image><image\_end><image\_start><image><image\_end><image\_start><image><image\_end><image\_start><image><image\_end>}
\end{tcolorbox}

Each task is designed to train the model's temporal understanding and visual reasoning capabilities.

\subsection{Potential Image Leakage in Testing Data}
When selecting data sources, we carefully choose those that do not overlap with the testing sets of our evaluation data, such as COCO~\citep{lin2014microsoft}. However, given that the data used in a \cref{sec: data} is composed of numerous sources, some degree of data leakage may be inevitable. As discussed and analyzed in a prior work~\citep{tong2024cambrian}, even when image overlap occurs, it does not necessarily imply that the exact image-question pairs have been encountered during training. Unlike traditional unimodal computer vision research, where an image alone constitutes a data point, the multimodal paradigm treats each image-text (question-answer) pair as a distinct and unique data point.

\section{Generating Visual Tokens}
Here, we include the quantitative results of all the experiments in \cref{sec: teach LLM to generate vision}.

\subsection{Results of Samples Needed to Unlock Visual Generation} \label{appendix: ablation on sample number}
\cref{tab: unlock vision} presents the quantitative results corresponding to \cref{fig: ablation on image-text}, which examines generation performance under two conditions: training exclusively on generation data and joint training with all other data described in \cref{sec: data}. The results demonstrate that the model can develop the ability for visual generation with a relatively modest amount of data when trained jointly with understanding tasks. In contrast, teaching this skill in isolation requires a substantially larger dataset.

In \cref{tab: data study generate vision}, we present the quantitative results corresponding to \cref{fig: data type comparison}, which investigates the impact of joint training on generation data in combination with various types of data outlined in \cref{sec: data}. The results show that joint training with visual understanding data—--specifically ImageQA and VideoQA--—provides the most significant improvement in visual generation performance.
\begin{table}[t]
    \centering
    \small 
    \begin{tabular}{cc|r}
    
      Joint train With Other Data & \# of Generation Data &
      FID Score \\
      \hline 

 Yes & 1k & 68.5  \\
    \rowcolor{Gray!10}No & 1k & 115.0  \\
 Yes & 5k & 19.2  \\
\rowcolor{Gray!10} No & 5k & 116,4  \\
 Yes & 10k & 18.7  \\
\rowcolor{Gray!10} No & 10k & 111.0  \\
 Yes & 50k & 17.1  \\
 \rowcolor{Gray!10}No & 50k & 111.8  \\
 Yes & 200k & 15.2  \\
\rowcolor{Gray!10} No & 200k & 110.7  \\
  Yes & 200k & 14.7  \\
 \rowcolor{Gray!10}No & 200k & 93.7  \\
  Yes & 1M & 14.4  \\
\rowcolor{Gray!10} No & 1M & 52.8  \\
  Yes & 3M & 15.1  \\
\rowcolor{Gray!10} No & 3M & 39.2  \\
  Yes & 5M & 14.3  \\
\rowcolor{Gray!10} No & 5M & 27.7  \\
    \end{tabular}.

\captionsetup{font={small}}
 \caption{\textbf{Results of training solely on generation data vs. joint training with additional data.} These results correspond to \cref{fig: ablation on image-text}. Joint training with additional data significantly improves generation performance. At 5,000 samples, the model begins to generate reasonably accurate visual tokens, indicating that visual generation is an ability unlocked through the learning of other tasks.}
\vspace{-0.25em}
\label{tab: unlock vision}
\end{table}
\begin{table}[t]
    \centering
    \small 
    \begin{tabular}{r|crr}
      Joint training Data & Data Type & FID Score &
      CLIP Score \\
      \hline 
 None & - & 110.5 & 5.7  \\
 Image-to-Image & Other Visual Data & 97.5 & 6.4  \\
 Visual Thinking & Other Visual Data & 93.5 & 6.5  \\
 Pure Video & Other Visual Data & 84.7 & 8.1  \\
 VideoQA & Visual Understanding Data & 26.5 & 16.1  \\
 ImageQA & Visual Understanding Data & 18.9 & 22.0  \\
 
    \end{tabular}.

\captionsetup{font={small}}
 \caption{\textbf{Impact of joint training 200k generation data with different data types.} These results correspond to \cref{fig: data type comparison}. Among the data types analyzed, joint training with visual understanding data has the most significant impact on enhancing visual generation performance.}
\vspace{-0.25em}
\label{tab: data study generate vision}
\end{table}

\subsection{Results of Joint training Different Understanding and Generation Data} \label{appendix: across data results}

In \cref{tab: 3*6 table}, we present the numerical results of joint training with varying scales of understanding data (1M, 4M, 7M) and generation data (200k, 500k, 1M, 2M, 3M, 4M). These findings demonstrate that increasing the amount of understanding data yields more substantial improvements in both understanding tasks (e.g., VQA performance) and generation tasks (e.g., FID scores and CLIP scores) compared to increasing the amount of generation data. These results, consistent with our analysis in \cref{sec: understanding for generation} and \cref{sec: asymmetric contribution}, highlight that understanding data play a more pivotal role in enhancing performance across both task types.

\begin{table*}[t]
    \centering
    \footnotesize 
    \setlength\tabcolsep{5.4pt} 
    \begin{tabular}{cc|cccccccccc|cc}
     \multicolumn{2}{c}{Data Composition} &
     \multicolumn{10}{c}{Image QA} &
     \multicolumn{2}{c}{Generation}  \\
     
      \# of VQA Data & \# of Generation Data &
    \rotatebox{90}{Average} &
      \rotatebox{90}{MMBench$^\text{EN}$} &
      \rotatebox{90}{SEED} &
      \rotatebox{90}{RealworldQA} &
      \rotatebox{90}{MMVP} &
      \rotatebox{90}{SQA} &
      \rotatebox{90}{MMMU} &
      \rotatebox{90}{VStar} &
      \rotatebox{90}{ChartQA} &
      \rotatebox{90}{TextVQA} &
      \rotatebox{90}{FID Score} &
      \rotatebox{90}{CLIP Score} \\
      \hline 
 1M & 200k & 46.4 & 60.0 & 62.2 & 50.3 & 24.0 & 80.0 & 38.4 & 37.4 & 16.4 & 48.8 & 28.3 & 15.2 \\
1M & 500k & 48.2 & 66.4 & 63.2 & 50.8 & 24.3 & 80.4 & 39.9 & 38.7 & 18.2 & 51.6 & 28.1 & 15.9 \\
1M & 1M & 49.1 & 70.1 & 65.2 & 52.2 & 21.3 & 80.0 & 39.5 & 38.7 & 20.4 & 54.6 & 27.3 & 16.5\\
1M & 2M & 49.9 & 67.8 & 66.0 & 50.2 & 30.3 & 80.2 & 38.9 & 39.0 & 21.8 & 54.8 & 23.1 & 17.8\\
1M & 3M & 51.1 & 71.3 & 67.1 & 55.4 & 33.0 & 79.5 & 38.8 & 37.4 & 22.7 & 55.0 & 21.1 & 21.1\\
1M & 4M & 51.4 & 71.1 & 66.9 & 52.4 & 31.0 & 80.5 & 39.8 & 41.1 & 24.0 & 56.0 & 18.4 & 22.3\\
          
\hdashline
4M & 200k & 53.8 & 73.1 & 68.8 & 55.0 & 34.7 & 81.2 & 38.5 & 44.0 & 29.5 & 59.2 & 21.4 & 20.5 \\
4M & 500k & 53.3 & 73.0 & 69.9 & 55.3 & 32.7 & 80.6 & 40.2 & 39.3 & 29.6 & 58.9 & 16.0 & 24.8\\
4M & 1M & 54.2 & 73.8 & 69.6 & 54.9 & 33.3 & 82.1 & 36.6 & 45.6 & 32.4 & 59.9 & 16.0 & 24.8\\
4M & 2M & 53.8 & 72.8 & 70.3 & 55.2 & 37.3 & 80.8 & 36.8 & 44.0 & 31.2 & 56.2 & 15.6 & 24.7\\
4M & 3M & 54.3 & 71.8 & 70.1 & 57.7 & 36.0 & 81.0 & 38.0 & 42.9 & 32.6 & 59.0 & 16.1 & 24.8\\
4M & 4M & 54.4 & 75.2 & 69.9 & 56.0 & 37.3 & 81.4 & 38.1 & 40.8 & 31.6 & 59.3 & 15.3 & 25.5\\

\hdashline
7M & 200k & 55.8 & 73.1 & 70.3 & 55.6 & 42.0 & 81.0 & 40.8 & 44.0 & 35.2 & 60.6 & 18.2 & 22.3\\
7M & 500k & 55.6 & 74.4 & 70.6 & 56.2 & 38.7 & 81.9 & 37.9 & 44.0 & 36.0 & 60.5 & 15.2 & 25.5\\
7M & 1M & 55.8 & 74.3 & 70.3 & 56.3 & 42.7 & 81.3 & 36.6 & 44.5 & 35.8 & 60.6 & 14.5 & 26.6\\
7M & 2M & 55.4 & 73.9 & 71.1 & 56.9 & 40.0 & 81.6 & 35.9 & 42.4 & 35.4 & 61.6 & 14.8 & 27.1\\
7M & 3M & 55.6 & 74.2 & 71.0 & 57.3 & 38.0 & 81.1 & 40.1 & 43.5 & 35.0 & 60.2 & 14.2 & 27.5\\
7M & 4M & 56.2 & 75.4 & 70.4 & 55.4 & 44.0 & 80.4 & 39.6 & 45.0 & 35.2 & 60.2 & 14.9 & 26.3\\

    \end{tabular}.

\captionsetup{font={small}}
 \caption{\textbf{Full results of joint training on varying amounts of VQA data (1M, 4M, 7M) and generation data (200k, 500k, 1M, 2M, 3M, 4M).} These results correspond to \cref{fig: vqa vs generation}, \cref{fig: generation vs vqa}, \cref{fig: comprehensive vqa vs gen}, and \cref{fig: vqa analysis}, which analyze how different combinations of understanding and generation data impact the model’s visual understanding and generation performance.}
\vspace{-0.25em}
\label{tab: 3*6 table}
\end{table*}

\subsection{Results of Training on Different LLMs} \label{appendix: across LLM results}
We present the results of training with 7M VQA data and 1M generation data across various LLM backbones, including LLaMA-3 8B, LLaMA-3.1 8B, and LLaMA-3 70B. As shown in \cref{tab: across LLM table}, which corresponds to the results in \cref{fig: ablation on LLMs}, we observe that stronger LLM backbones lead to improvements in both visual understanding and visual generation. These findings further support the conclusion that visual understanding and generation are reciprocal processes, where advancements in one drives enhancements in the other.

\begin{table*}[t]
    \centering
    \small 
    \begin{tabular}{c|cccccccccc|cc}
     \multicolumn{1}{c}{Pretrained LLM} &
     \multicolumn{10}{c}{Image QA} &
     \multicolumn{2}{c}{Generation}  \\     
      LLM &
    \rotatebox{90}{Average} &
      \rotatebox{90}{MMBench$^\text{EN}$} &
      \rotatebox{90}{SEED} &
      \rotatebox{90}{RealworldQA} &
      \rotatebox{90}{MMVP} &
      \rotatebox{90}{SQA} &
      \rotatebox{90}{MMMU} &
      \rotatebox{90}{VStar} &
      \rotatebox{90}{ChartQA} &
      \rotatebox{90}{TextVQA} &
      \rotatebox{90}{FID Score} &
      \rotatebox{90}{CLIP Score} \\
      \hline

LLaMA-3 8B & 55.8 & 74.3 & 70.3 & 56.3 & 42.7 & 81.3 & 36.6 & 44.5 & 35.8 & 60.6 & 14.5 & 26.6
\\
LLaMA-3.1 8B & 56.7 & 75.8 & 70.2 & 56.2 & 44.7 & 81.9 & 41.2 & 43.4 & 36.0 & 61.3 & 13.2 & 27.1
\\
LLaMA-3 70B & 60.7 & 80.7 & 72.6 & 58.3 & 48.7 & 87.8 & 48.9 & 47.1 & 37.4 & 65.0 & 13.8 & 26.8
\\

    \end{tabular}.

\captionsetup{font={small}}
 \caption{\textbf{Full results of training on different LLMs.} We train 7M VQA data and 1M generation data on different LLM backbones (LLaMA-3 8B, LLaMA-3.1 8B, and LLaMA-3 70B) and measure understanding and generation performance.  }
\vspace{-0.25em}
\label{tab: across LLM table}
\end{table*}

\section{More Examples of \ours{}}
\begin{figure*}[t]
    \vspace{-1.25em}
    \centering
    \includegraphics[width=\linewidth]{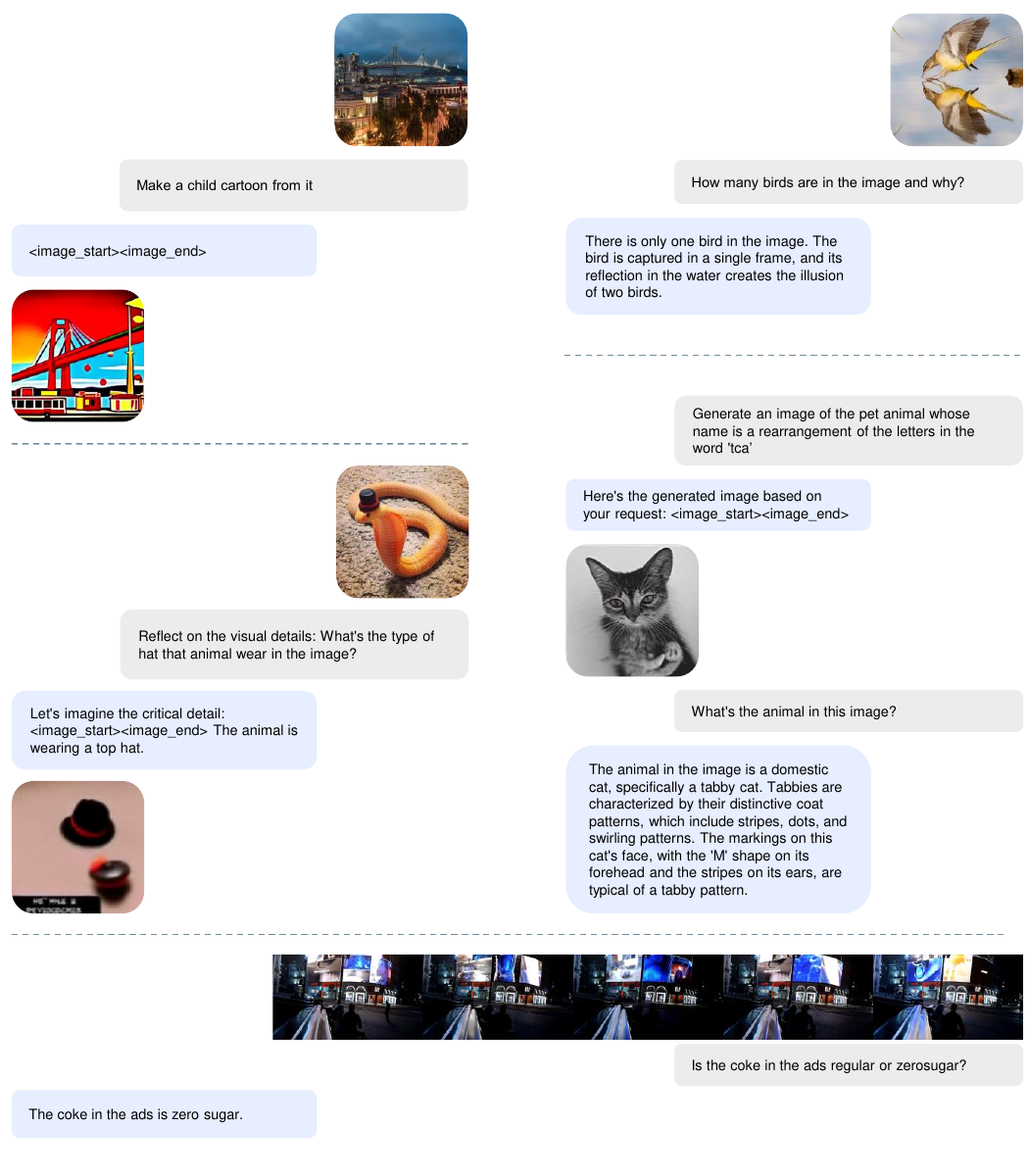}
\vspace{-2em}
    \caption{\small \textbf{Examples of \ours{} (I).} We showcase examples of \ours{}'s capabilities: transforming images based on prompts (\textit{top-left}), answering challenging questions (\textit{top-right}), integrating visual tokens into reasoning processes (\textit{bottom-left}), implicitly solving puzzles (\textit{bottom-right}) and answering tricky video-qa questions. (\textit{bottom}). }
    \label{fig: metamorph examples I}
    \vspace{-1.25em}
\end{figure*}

\begin{figure*}[t]
    \vspace{-1.25em}
    \centering
    \includegraphics[width=\linewidth]{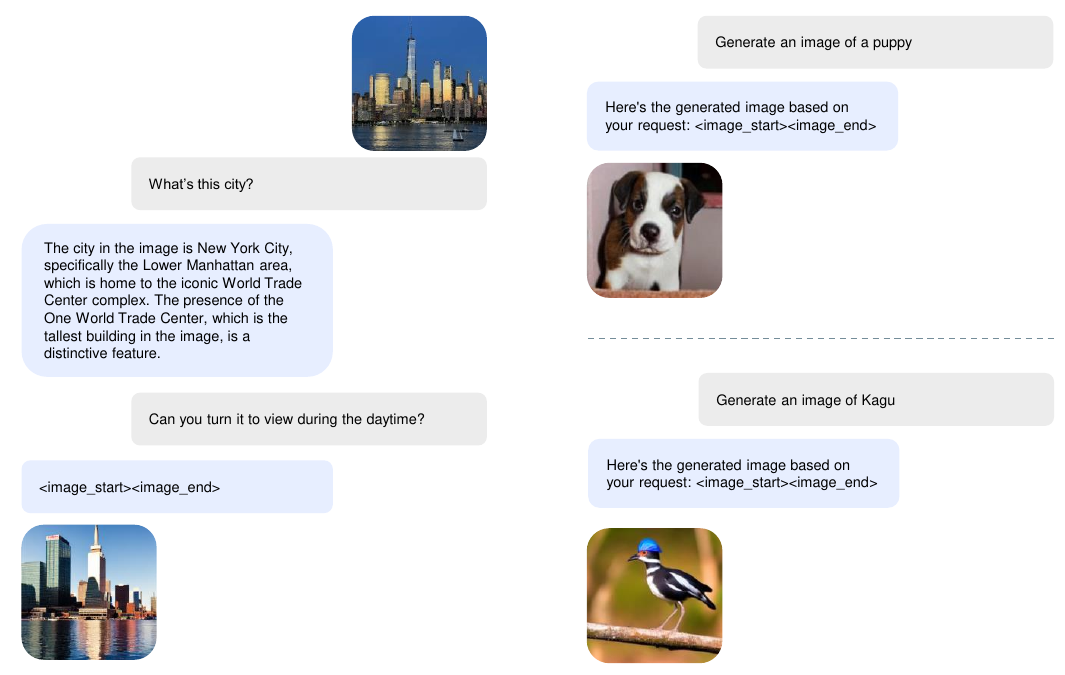}
\vspace{-2em}
    \caption{\small \textbf{Examples of \ours{} (II).}We showcase more examples of \ours{}'s capabilities: answering questions and transforming images in one conversation (\textit{left}), generating images (\textit{top-right}), and leveraging knowledge in LLMs to generate rare concepts (\textit{bottom-right}). }
    \label{fig: metamorph examples II}
    \vspace{-1.25em}
\end{figure*}

We provide additional examples of \ours{} in \cref{fig: metamorph examples I} and \cref{fig: metamorph examples II}. These examples illustrate how \ours{} extends beyond the capabilities of typical MLLMs by leveraging learned skills to perform novel tasks such as visual reasoning and visual transformation. In \cref{fig: metamorph examples I}, when prompted with the question \textit{``What is the type of hat?''}, \ours{} first generates visual tokens related to hats and then answers correctly with \textit{``top hat''}. The model also demonstrates the ability to perform image transformations, such as creating a cartoon version of an image or altering it to appear as daytime. Additionally, we showcase examples of \ours{} solving implicit puzzles, such as interpreting \textit{``a rearrangement of the letters in the word 'tca'''}, before generating the corresponding visual tokens of cats. 

\end{document}